\pdfoutput=1
\documentclass{article} 
\let\StandardAddContentsLine\addcontentsline
\usepackage{iclr2027_conference,times}

\usepackage{amsmath,amsfonts,bm}

\def\eqref#1{equation~\ref{#1}}

\def\1{\bm{1}}

\DeclareMathAlphabet{\mathsfit}{\encodingdefault}{\sfdefault}{m}{sl}
\SetMathAlphabet{\mathsfit}{bold}{\encodingdefault}{\sfdefault}{bx}{n}

\usepackage{amsmath,amssymb,amsfonts,mathtools}
\usepackage{graphicx}
\usepackage{booktabs}
\usepackage{longtable}
\usepackage{multirow}
\usepackage[table]{xcolor}
\usepackage{array,etoolbox,flafter}
\usepackage{url}
\usepackage{etoc}
\usepackage{microtype}

\definecolor{TableBand}{RGB}{240,242,244}
\definecolor{TableCI}{RGB}{125,135,145}
\definecolor{PaperInk}{HTML}{26323F}
\definecolor{PaperRule}{HTML}{AAB4C1}
\definecolor{FigBlue}{RGB}{61,105,150}
\definecolor{FigTeal}{RGB}{34,133,117}
\definecolor{FigGreen}{RGB}{39,131,116}
\definecolor{FigOrange}{RGB}{198,122,63}
\definecolor{FigGray}{RGB}{171,184,199}
\definecolor{FigDark}{RGB}{42,58,72}
\definecolor{FigGrid}{RGB}{224,230,236}
\definecolor{StrictColor}{RGB}{66,94,124}
\definecolor{MVColor}{RGB}{95,99,104}
\definecolor{HeatBlue}{RGB}{82,118,155}
\definecolor{GridGray}{RGB}{222,225,228}

\AtBeginEnvironment{table}{\renewcommand{\arraystretch}{1.10}\arrayrulecolor{PaperRule}}
\AtBeginEnvironment{longtable}{\renewcommand{\arraystretch}{1.10}\arrayrulecolor{PaperRule}}

\makeatletter
\@ifpackagelater{longtable}{2025/12/23}{}{%
  \patchcmd{\LT@output}{\vss}{\vskip\z@\@plus1fil\@minus\normalbaselineskip}{}{}%
  \patchcmd{\LT@output}{\vss}{\vskip\z@\@plus1fil\@minus\normalbaselineskip}{}{}%
}
\makeatother

\newcommand{\hash}{\texttt{\#\#\#\#}}
\newcommand{\boxedm}{\texttt{\textbackslash boxed\{\}}}
\let\addcontentsline\StandardAddContentsLine
\usepackage[hidelinks,linktoc=all,bookmarksdepth=2]{hyperref}
\let\AppendixAddContentsLine\addcontentsline
\renewcommand{\addcontentsline}[3]{}

\title{Learned Reporting Preferences in RLVR\\Can Conflict with the Current Request}

\author{%
   Yupeng Chang\textsuperscript{1}, Wenxuan Zhang\textsuperscript{3}, Yuan Wu\textsuperscript{1,2}\thanks{Corresponding author} \\
   \textsuperscript{1}School of Artificial Intelligence, Jilin University\\
   \textsuperscript{2}Key Laboratory of Symbolic Computation and Knowledge Engineering, Jilin University\\
   \textsuperscript{3}Singapore University of Technology and Design\\
   {\small\texttt{changyp23@mails.jlu.edu.cn, wxzhang@sutd.edu.sg, yuanwu@jlu.edu.cn}} \\
}

\iclrfinalcopy 
\begin{document}

\maketitle
\fancyhead{} 

\begin{abstract}
Reinforcement learning with verifiable rewards (RLVR) has become a prominent
approach for improving language-model performance on reasoning tasks using
automatically checked answers.
Yet convention-matched evaluation cannot reveal whether reinforcing one
reporting convention reduces adherence to a different request that the initial
policy already follows.
To test this, we train matched policies under two reporting conventions and
evaluate each policy under both current requests, using the same initial policy
as a shared reference.
We complement this crossed design with controlled interventions and independent
human calibration.
On GSM8K, boxed-format RLVR reduces the fraction of Qwen2.5-7B responses
containing the requested hash-format payload by
35.33--74.37 percentage points relative to a 95.45\% initial baseline in four
of five training seeds; the fifth improves by 2.50 points.
In the four deteriorating runs, almost every response that omits the requested
payload instead retains the trained boxed convention, and the
same four seeds deteriorate under two fixed paraphrases.
Changing only the final-answer marker in supervised targets reverses which
reporting convention the model prefers across three seeds, providing controlled
evidence that this preference is learnable.
Across three settings with independent human calibration, gains under a
convention-sensitive scorer exceed the corresponding gains in
committed-answer correctness, i.e., the correctness of the answer the model
actually commits to.
Together, these results separate three distinct post-training outcomes:
learned reporting preference, current-request adherence, and committed-answer
correctness.
They show that convention-matched accuracy alone does not fully characterize
post-training behavior and motivate evaluating current-request adherence
alongside convention-matched task accuracy.
\textbf{Code:} \href{https://github.com/llm172/rlvr-reporting-conventions}{\textcolor{blue!65!black}{GitHub repository}}.
\end{abstract}

\etocdepthtag.toc{chapter}
\section{Introduction}

Reinforcement learning with verifiable rewards (RLVR) trains language models
on reasoning tasks using automatically checked answers
\citep{shao2024deepseekmath,guo2025deepseek,lambert2024tulu}.
In practice, however, these rewards often couple answer correctness to a
machine-readable reporting convention.
As a result, training can reinforce not only \emph{what} answer a model
produces, but also \emph{how} it reports that answer.
This raises a reliability question that convention-matched evaluation does not
answer:
\emph{when RLVR reinforces one reporting convention, does the model still
follow a different request that the initial policy already followed?}

Consider a model trained to place its final answer inside
\(\boxed{\cdot}\), but later asked to report the answer after the literal
four-hash delimiter \hash{}.
Both conventions can represent the same answer, yet a learned reporting
preference may compete with the current request.
The model may produce an answer elsewhere in the response while failing
to place it in the format explicitly requested by an evaluator, tool, or
downstream interface.
Such a regression can coexist with improved performance under the rewarded
convention and remain invisible to convention-matched evaluation.
Figure~\ref{fig:reporting-baseline} illustrates this behavior in our
experiments.

Prior work establishes several phenomena:
RLVR can correct prompt--format misalignment \citep{wu2025invisible}, verifier
failures can miss correct responses or provide misleading reward signals
\citep{huang2025accuracy}, reasoning-oriented post-training can degrade instruction
following \citep{fu2026scaling}, and RLVR gains need not reduce to format
correction alone \citep{wang2026reinforcement}.
These findings show that reporting format, verification, and task performance
can interact.
However, they do not directly answer whether training under one reporting
convention makes the model worse at following another request that it could
follow before training.
Convention-matched evaluation cannot reveal this conflict.
Changing only the evaluation prompt can reveal prompt sensitivity, but it does
not show how training changed behavior relative to the initial policy.
Likewise, agreement among automatic readers does not establish that a score
gain reflects an improvement in the answer the model actually commits to.

We address this gap with a crossed
\emph{training-convention $\times$ current-request} evaluation.
Within each seed, we train matched policies under two reporting conventions and
evaluate each policy under both requests.
This design separates the convention reinforced during training from the
request given at evaluation time.
The same initial policy serves as a common reference, allowing us to measure
whether training improves or degrades behavior that was already present before
RLVR.
We measure whether the requested payload is present regardless of correctness
and rescore the same responses with fixed readers.
In a controlled SFT intervention, we keep the target content fixed and change
only the final-answer marker to test whether reporting preference itself is
learnable.
Independent human calibration in selected settings tests whether automatic
score gains correspond to gains in committed-answer correctness.
Together, these measurements separate three outcomes that convention-matched
accuracy can conflate:
\emph{learned reporting preference, current-request adherence, and
committed-answer correctness.}

The experiments reveal a substantial but non-universal loss of
current-request adherence.
On GSM8K, after boxed-format RLVR, the fraction of Qwen2.5-7B responses
containing the requested hash-format payload falls by
35.33--74.37 percentage points from a 95.45\% initial baseline in four of five
training seeds; the fifth improves by 2.50 points.
In the four deteriorating runs, almost every response that omits the requested
payload instead retains the trained boxed convention.
The same four seeds also deteriorate under two fixed paraphrases.
Changing only the final-answer marker in supervised targets reverses which
reporting convention the model prefers across three seeds, providing controlled
evidence that this preference is learnable.
Across three settings with independent human calibration, gains under a
convention-sensitive scorer exceed the corresponding gains in
committed-answer correctness.
These comparisons do not rule out semantic improvement.

These results show that convention-matched accuracy captures only part of
post-training behavior.
A policy can improve under the interface used during training while becoming
less responsive to another interface it previously handled successfully.
Current-request adherence is therefore a distinct evaluation target alongside
convention-matched task accuracy and committed-answer correctness.

Our contributions are threefold:
\begin{itemize}
    \item We identify \emph{loss of previously supported request adherence} as
    a distinct RLVR failure mode that convention-matched evaluation can miss.

    \item We introduce a crossed
    \emph{training-convention $\times$ current-request} protocol, anchored to a
    shared initial policy, that separates training-acquired reporting
    preference from current-request adherence and reader-dependent scoring.

    \item Using five paired training seeds, fixed paraphrases, controlled SFT,
    multiple readers, and independent human calibration, we show that learned
    reporting preference, current-request adherence, and committed-answer
    correctness need not move together after post-training.
\end{itemize}
\section{Measurement Framework and Experimental Design}
\label{sec:decomp}

Convention-matched evaluation can conflate four distinct aspects:
the convention reinforced during training, the convention requested at
evaluation, the reader used to extract an answer, and the correctness of the
answer the response commits to.
To separate these sources of variation, we cross
\emph{training convention} with \emph{current request}, use the same initial
policy as a common reference, and compare requested-payload behavior,
fixed-reader scores, and human judgments.
These measurements distinguish three post-training outcomes:
learned reporting preference, current-request adherence, and
committed-answer correctness.

\subsection{Crossing training convention with current request}
\label{sec:pair}

Let $x\sim\mathcal D$ denote a problem with reference answer
$a^\star(x)\in\mathcal A$.
We train matched policies $\pi_t$ under reporting convention
$t\in\{H,B\}$ (hash/boxed), and evaluate each policy under request
$r\in\{h,b\}$:
\[
y\sim\pi_t(\cdot\mid p_r(x)).
\]
Changing $t$ changes the training prompt and matched reward extractor while
holding the remaining training configuration fixed.
Changing $r$ keeps model weights fixed but changes the evaluation prompt and
response distribution.
The same initial policy $\pi_0$ is evaluated under both requests and serves as
a common reference, separating behavior acquired during training from
request-following behavior present before RLVR.
This crossed design therefore asks a question that convention-matched
evaluation cannot: whether training under one reporting interface changes how
well the model follows another current request.

\paragraph{Current-request adherence.}
Let $d_r(y)$ indicate whether a response contains a payload in the convention
explicitly requested by the evaluation prompt, independently of correctness.
Define
\[
P_{t,r}=\mathbb E[d_r(y)],\qquad
\Delta P_{t,r}=P_{t,r}-P_{0,r},
\]
with $P_{0,r}$ defined analogously for the initial policy.
Thus, $\Delta P_{t,r}<0$ indicates deterioration relative to behavior
supported by the initial policy.
We summarize the training-by-request interaction as
\begin{equation}
I_P=
100\Big[
(P_{H,h}-P_{B,h})-(P_{H,b}-P_{B,b})
\Big].
\label{eq:payloadinteraction}
\end{equation}
$I_P$ is the hash-trained policy's payload advantage under the hash request
minus its advantage under the boxed request, in percentage points.
It measures reporting behavior, not answer correctness or latent capability.
For GSM8K, the hash convention uses the literal four-hash delimiter \hash{}.
The detector accepts a numeric payload after this delimiter on the same line or
nonempty content inside balanced \boxedm{} braces; any qualifying occurrence
suffices, including responses containing both formats.
Thus, $d_r$ measures requested-format presence, not final commitment, value
correctness, or agreement.
Full rules appear in Appendix~\ref{app:reporting-baseline}.

\subsection{Reader-dependent accuracy}
\label{sec:identity}

Requested-payload presence and answer accuracy are distinct properties of the
same response.
Let $E:\mathcal Y\to\mathcal A\cup\{\bot\}$ be a fixed reader that returns a
normalized answer or extraction failure:
\begin{equation}
r_E(x,y)=\mathbf 1[E(y)=a^\star(x)],\qquad
\mathrm{Acc}(\pi,p,E)=\mathbb E_{x,y}[r_E(x,y)].
\label{eq:acc}
\end{equation}
When the prompt is fixed, we write
$\mathrm{Acc}_E(\pi)=\mathrm{Acc}(\pi,p,E)$.

For a convention-specific reader $E_c$, let
$q_c=\Pr[E_c(y)\ne\bot]$ denote extractability and
$u_c=\Pr[E_c(y)=a^\star(x)\mid E_c(y)\ne\bot]$ denote conditional
correctness.
When $q_c>0$,
\begin{equation}
\mathrm{Acc}(\pi,p,E_c)=q_cu_c.
\label{eq:gate}
\end{equation}
This decomposition is descriptive, not causal:
score changes may reflect extractability, conditional correctness, or both.
In symbolic transfer, requested-marker payload presence $q_c^{\rm pay}$ is
measured before parsing and need not equal $q_c$.

Readers remain fixed within comparisons and need not match the training reward
program.
Let $A_{t,r}^{(R)}$ denote crossed-cell accuracy under reader $R$.
The \emph{strict} reader follows the requested convention, whereas
Math-Verify (MV) uses the same full-response reader across requests.
We define the corresponding reader-specific interaction as
\begin{equation}
I_R=
\big(A_{H,h}^{(R)}-A_{B,h}^{(R)}\big)
-
\big(A_{H,b}^{(R)}-A_{B,b}^{(R)}\big).
\label{eq:crossinteraction}
\end{equation}
We report $100I_R$ in percentage points.
Unlike $I_P$, which measures payload behavior, $I_R$ measures
reader-specific accuracy.

\paragraph{Initial-to-RL gain accounting.}
For the initial policy $\pi_0$ and an RL descendant $\pi_{\rm RL}$, define
\[
\gamma_m=
\mathrm{Acc}_{E_{\rm len}}(\pi_m)-
\mathrm{Acc}_{E_{\rm str}}(\pi_m),
\qquad m\in\{0,{\rm RL}\},
\]
and
\[
\Delta_S
=
\mathrm{Acc}_{E_{\rm str}}(\pi_{\rm RL})
-
\mathrm{Acc}_{E_{\rm str}}(\pi_0),
\qquad
\Delta_{\rm len}
=
\mathrm{Acc}_{E_{\rm len}}(\pi_{\rm RL})
-
\mathrm{Acc}_{E_{\rm len}}(\pi_0).
\]
Then
\begin{equation}
\Delta_S
=
\underbrace{(\gamma_0-\gamma_{\rm RL})}_{F:\ \text{reader-gap reduction}}
+\Delta_{\rm len}.
\label{eq:decomp}
\end{equation}
This is an accounting identity, not a causal decomposition of reasoning.

Under the stronger fidelity condition
\begingroup
\renewcommand{\theHequation}{assumption-a1}
\begin{equation}
E_{\rm len}(y)=E_{\rm str}(y)
\qquad\text{whenever }E_{\rm str}(y)\ne\bot,
\tag{A1}\label{eq:a1}
\end{equation}
\endgroup
the reader gap corresponds to credit that the lenient reader can recover when
the strict reader fails.
The hash-first/fallback reader satisfies A1 by construction, although
a recovered value need not be the response's committed answer.

\subsection{Human calibration of committed-answer correctness}

Automatic readers recover extractable answers, but do not necessarily identify
the answer a response commits to.
We therefore use independent human calibration in selected settings.

Let $Z(x,y)\in\{0,1\}$ indicate committed-answer correctness and
$H_m=\Pr_{\pi_m}[Z=1]$.
For reader $E$, define false credit and missed correct credit as
\[
e_{E,m}^{+}=\Pr[r_E=1,Z=0],\qquad
e_{E,m}^{-}=\Pr[r_E=0,Z=1],
\]
and define net reader bias as
\[
b_{E,m}=e_{E,m}^{+}-e_{E,m}^{-}.
\]

The corresponding human and reader gains are
\[
\Delta_H=H_{\rm RL}-H_0,\qquad
\Delta_E=
\mathrm{Acc}_E(\pi_{\rm RL})-\mathrm{Acc}_E(\pi_0).
\]
Then
\begin{equation}
\mathrm{Acc}_E(\pi_m)=H_m+b_{E,m},\qquad
\Delta_E-\Delta_H=b_{E,{\rm RL}}-b_{E,0}.
\label{eq:readerbias}
\end{equation}
Thus, the reader--human gain gap measures the change in net reader bias from
the initial policy to its RL descendant.
We leave ambiguous or conflicting commitments unresolved.
Agreement among automatic readers cannot establish committed-answer
correctness.
$Z=1$ certifies committed-answer correctness, not derivation validity.

\subsection{Controlled interventions and boundary analyses}
\label{sec:design}

We use four interventions to separate acquisition from evaluation effects.
The crossed design varies
\textbf{I2: training convention} and
\textbf{I3: current request}.
Two complementary interventions test how reporting preference is acquired:
\textbf{I1: reward requirement} removes the marker requirement while retaining
answer correctness, and
\textbf{I4: supervised marker} changes only the final marker in content-matched
SFT targets, with an unchanged-marker control.

We summarize these controls using gain attenuation and prompted convention
preference:
\begin{align}
\rho
&=
1-
\frac{\Delta_S(\mathrm{treatment})}
     {\Delta_S(\mathrm{control})},
&&\text{gain attenuation},
\label{eq:rho}\\
D(\pi)
&=
\mathrm{Acc}(\pi,p_c,E_c)
-
\mathrm{Acc}(\pi,p_{c'},E_{c'}),
&&\text{prompted convention preference},
\label{eq:dpref}
\end{align}
where $c\ne c'$ denote reporting conventions.
Gains use each arm's initial score.
When the control gain is positive, $\rho>1$ means that the treatment gain is
negative.
Because $D$ pairs each request with its corresponding reader, it measures
prompted convention preference rather than instruction following independently
of the reader.
For I4, we use the arm-to-arm contrast, written relative to the shared initial policy as
\[
[D(\pi_H^{\rm SFT})-D(\pi_0)]
-
[D(\pi_B^{\rm SFT})-D(\pi_0)],
\]
where $H/B$ denote the hash-marker and unchanged boxed-marker SFT arms.

\paragraph{Finite-budget coverage.}
As a boundary analysis, we also ask whether at least one correct answer is
observed within a fixed sampling budget.
For $k$ independently sampled responses per item, define
\begin{equation}
S_k(\pi,E)=
\left\{
x:\exists\,i\le k,\;
E(y_i)=a^\star(x),\;
y_i\sim\pi(\cdot\mid p(x))
\right\}.
\label{eq:support}
\end{equation}
This random, reader-dependent set measures observed finite-budget coverage,
not full policy support.
Estimator details and paired tests appear in
Appendix~\ref{app:coverage-estimator}.
Nonsignificant reader comparisons establish neither loss nor equivalence.
\section{Experiments}

\label{sec:setup}

\paragraph{Models, tasks, and matched controls.}
Our main benchmark is the $1{,}319$-item GSM8K test split
\citep{cobbe2021training}.
We study Qwen2.5-Instruct at 1.5B and 7B
\citep{qwen2025qwen2}, SmolLM2-1.7B-Instruct
\citep{allal2025smollm2}, and OLMo-2-7B-Instruct
\citep{olmo20242}.
Unless noted, training uses $100$ GRPO steps \citep{shao2024deepseekmath}, batch size $32$,
and $8$ rollouts per prompt.
Paired intervention arms share the backbone, frozen training data, ordering,
and configuration except for the specified intervention.
MATH500 \citep{hendrycks2021measuring} evaluates native-\boxedm{} training and
crossed-request transfer of GSM8K-trained 7B checkpoints, while code
benchmarks test transfer from math training.
For 7B crossed training, both arms use the frozen stratified prompt
proposal with importance correction
(Appendix~\ref{app:train-config}).
The GSM8K crossed comparison uses five paired training seeds ($83$--$87$).
Seeds $86$ and $87$ were fixed before the extension, while seeds $83$--$85$
were retained; all five are reported without outcome-based exclusions.
A wording check evaluates the same checkpoints on a fixed $300$-item subset
under two additional request paraphrases.

\paragraph{Matched evaluation and readers.}
Table~\ref{tab:matched} evaluates the initial and step-$100$ RL checkpoints on all
$1{,}319$ GSM8K items using identical prompts and item seeds, with $T=0.6$,
top-$p=0.95$, a $640$-token output budget, and a $4096$-token context,
without input truncation.
For Qwen2.5-1.5B and SmolLM2, we additionally evaluate fixed rephrased and
one-example prompts and repeat the matrix with greedy decoding.
The strict reader returns the first parseable answer after \hash{}; the
lenient reader preserves that answer when present and otherwise falls back to
the last number.
Last-\hash{}, unconditional-last-number, and Math-Verify (MV)\footnote{\url{https://github.com/huggingface/Math-Verify}} provide three
additional readers over the same saved responses
(Appendix~\ref{app:matched}).
A fresh $500$-item arithmetic probe uses an AST/Fraction checker and disjoint
development fixtures.
Greedy and $32$-sample readouts use no checkpoint selection.
The $7{,}377$ GSM8K training questions have zero normalized-question overlap
with the $1{,}319$-item evaluation split; pretraining overlap is unknown.

\paragraph{Transfer evaluation.}
The 7B MATH500 transfer panel evaluates all $500$ items under both requests,
with one response per condition, using $T=0.6$, top-$p=0.95$, and an
$8192$-token output budget.
Frozen symbolic readers score the requested hash or boxed payload, while
full-response MV parses independently; the training reward uses a
convention-specific program
(Appendices~\ref{app:math-transfer} and~\ref{app:train-config}).
Hash-reward training is repeated for seeds $84/85$ for $100$ updates; the
initial and seed-$83$ responses are reused
(Appendix~\ref{app:qwen7-seeds}).

\paragraph{Uncertainty and scope.}
Automatic-score intervals use paired item bootstraps with $2000$ resamples,
or $20000$ for transfer and hash-only replication, with sampled
item indices shared across matched cells.
These intervals condition on fixed checkpoints and saved responses, do not
quantify training-seed uncertainty, and are not multiplicity-corrected.
Five-seed GSM8K and wording summaries instead report equally weighted means,
sample SDs across seeds, and individual outcomes; they describe the fixed seed
set rather than a training-seed population.
Human calibration uses stratified finite-population bounds with simultaneous
correction and unresolved-label sensitivity
(Appendix~\ref{app:human-calibration}).
New protocols used for the reported prospective controls were frozen before generation.

\begin{table}[!htb]
\caption{\textbf{Matched GSM8K scores and gains across models, prompt variants,
and readers.}
Scores are percentages; changes are percentage points.}
\label{tab:matched}

\centering
\footnotesize
\setlength{\tabcolsep}{3.8pt}
\renewcommand{\arraystretch}{1.08}

\begin{tabular*}{0.98\linewidth}{
@{\extracolsep{\fill}}
llrrrrc
@{}}
\toprule

Model / seed
& Prompt
& $S_0$
& $S_{\rm RL}$
& $F$
& $\Delta L$
& \shortstack{$\Delta$MV\\[-1pt]{\scriptsize 95\% CI}} \\
\midrule

\rowcolor{TableBand}
\multicolumn{2}{@{}l}{\textbf{A. Default prompt}}
&
\multicolumn{5}{r@{}}{\itshape Matched comparison across model families} \\

Qwen2.5 1.5B / 83
& Default
& 11.1
& 75.4
& +59.7
& +4.6
& $+3.9$ {\color{TableCI}$[+1.4,+6.4]$} \\

SmolLM2 1.7B / 83
& Default
& 10.2
& 41.5
& +26.2
& +5.2
& $+4.1$ {\color{TableCI}$[+1.6,+6.6]$} \\

Qwen2.5 7B / 83
& Default
& 77.7
& 91.6
& +14.1
& -0.2
& $+0.0$ {\color{TableCI}$[-1.2,+1.3]$} \\

OLMo-2 7B / 83
& Default
& 65.6
& 85.4
& +15.8
& +4.1
& $+3.8$ {\color{TableCI}$[+1.8,+5.7]$} \\

OLMo-2 7B / 84
& Default
& 65.6
& 85.1
& +15.8
& +3.7
& $+3.5$ {\color{TableCI}$[+1.5,+5.5]$} \\

OLMo-2 7B / 85
& Default
& 65.6
& 86.1
& +15.8
& +4.7
& $+4.4$ {\color{TableCI}$[+2.5,+6.4]$} \\

\addlinespace[2pt]

\rowcolor{TableBand}
\multicolumn{2}{@{}l}{\textbf{B. Prompt variants}}
&
\multicolumn{5}{r@{}}{\itshape Same variant applied to both checkpoints} \\

Qwen2.5 1.5B / 83
& Rephrased
& 54.4
& 74.5
& +16.1
& +3.9
& $+4.1$ {\color{TableCI}$[+1.7,+6.4]$} \\

Qwen2.5 1.5B / 83
& One-shot
& 55.5
& 72.3
& +13.1
& +3.6
& $+4.1$ {\color{TableCI}$[+1.7,+6.5]$} \\

SmolLM2 1.7B / 83
& Rephrased
& 36.9
& 40.3
& +2.4
& +1.0
& $+0.9$ {\color{TableCI}$[-1.2,+3.0]$} \\

SmolLM2 1.7B / 83
& One-shot
& 39.7
& 43.5
& +2.4
& +1.4
& $+1.5$ {\color{TableCI}$[-0.5,+3.6]$} \\

\bottomrule
\end{tabular*}

\vspace{2pt}
\parbox{0.97\linewidth}{\scriptsize
GSM8K, $1{,}319$ items, $T=0.6$. Initial and RL prompts, decoding, and item
seeds are matched within each row. $S$ is first-hash strict, $L$ is
hash-first/fallback, and MV is Math-Verify. $F$ is strict--lenient
reader-gap reduction, with $\Delta S=F+\Delta L$.
Brackets report paired 95\% item-bootstrap intervals for $\Delta$MV,
conditional on fixed checkpoints and saved responses.
Minor discrepancies reflect rounding.}
\end{table}

\begin{figure}[!htb]
\centering
\includegraphics[width=.98\textwidth]{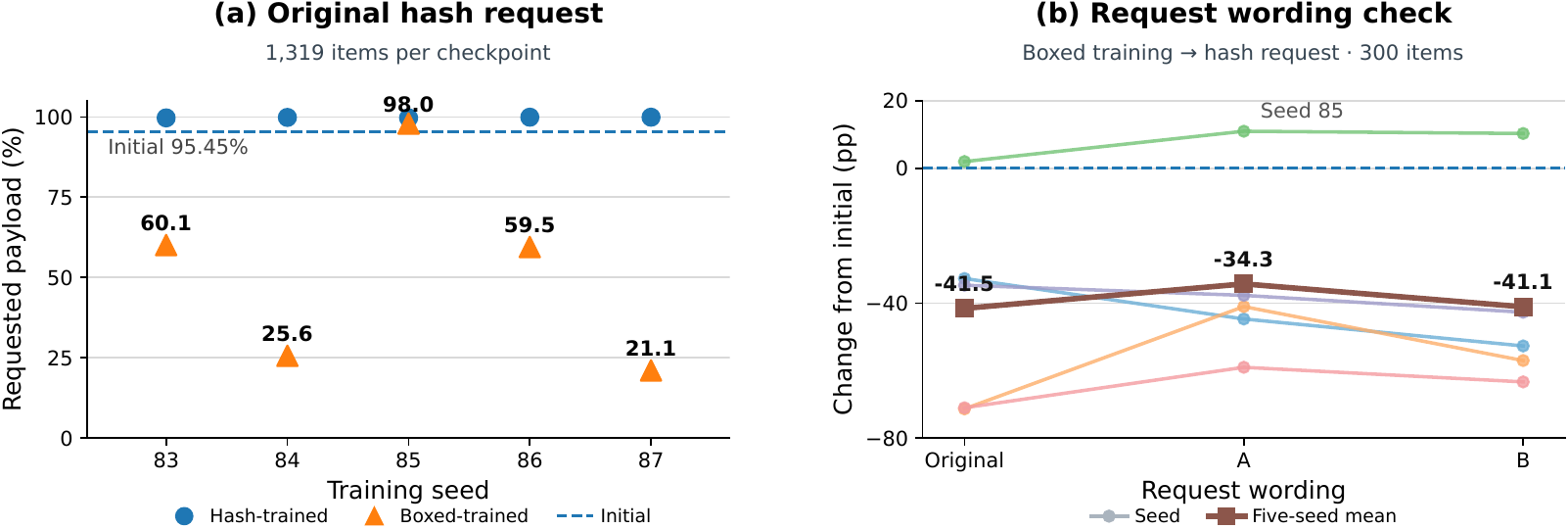}
\caption{\textbf{Current-request adherence across seeds and request wordings.}
\textbf{(a)} Requested hash-payload presence, irrespective of correctness,
on all $1{,}319$ GSM8K items; the dashed line marks the shared initial policy.
\textbf{(b)} Change from each wording's initial-policy baseline after boxed
training on a fixed $300$-item subset using the original request and two
paraphrases (A/B). Thin lines connect the same five seeds; thick lines show
their equally weighted means. Four seeds deteriorate under all three wordings,
while seed $85$ improves under all three. Wordings are repeated evaluations,
not additional training seeds.
Details appear in Appendices~\ref{app:reporting-baseline}--\ref{app:wording-check}.}
\label{fig:reporting-baseline}
\end{figure}

\subsection{RQ1: Can a learned reporting preference interfere with current-request adherence?}
\label{sec:causal}

Crossed evaluation tests current-request adherence, while content-matched
supervision and reward/convention controls probe reporting-preference
acquisition.

\paragraph{Current-request adherence can deteriorate while the trained convention persists.}
A fixed detector measures requested-format payload presence independently of
correctness across all $29{,}018$ initial and trained responses
(Figure~\ref{fig:reporting-baseline}a).
From an initial hash-payload rate of $95.45\%$, boxed training changes
presence by $-35.33$, $-69.83$, $+2.50$, $-35.94$, and $-74.37$
percentage points for seeds $83$--$87$.
Both extension seeds ($86$ and $87$), fixed beforehand, also deteriorate, while
the seed-$85$ improvement is retained.
Under hash requests, hash-trained policies reach at least $99.70\%$
payload presence; under boxed requests, all trained policies reach at least
$99.55\%$.
Responses containing both formats count as providing the requested payload,
without implying correctness.
Under hash requests, boxed-only responses account for almost all omissions
in the four deteriorating boxed-trained runs
(Table~\ref{tab:reporting-payload-counts}); only $31/1{,}319$ initial responses
contain a box without a hash payload.
Thus, responses that omit the requested payload typically retain the trained
convention, supporting an output-level reporting preference.
Exact counts and additional reader/item-level checks appear in
Appendix~\ref{app:reporting-baseline}.

The pattern persists under wording changes.
On a fixed $300$-item subset, we evaluate the original request and two
paraphrases for all five seed pairs.
Relative to each initial-policy baseline, boxed training reduces hash-payload
presence in the same four seeds under every wording, while seed $85$ improves
under all three (Figure~\ref{fig:reporting-baseline}b).
The five-seed mean changes are $-41.53$, $-34.27$, and $-41.07$ percentage
points.
Thus, the effect is not confined to the original wording, although its
magnitude and occurrence remain seed dependent.
The three wordings reuse the same checkpoints and are not independent
training repetitions (Appendix~\ref{app:wording-check}).

\begin{figure*}[t]
    \centering
    \includegraphics[width=\textwidth]{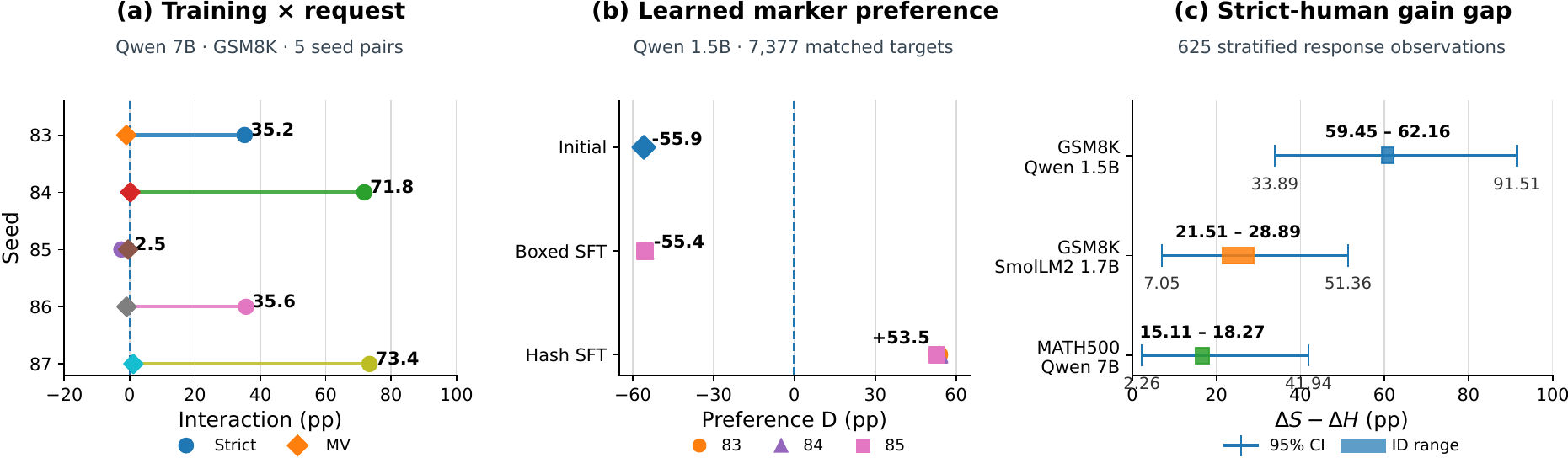}
\caption{
\textbf{Complementary evidence separates reporting behavior, learned
preference, and committed-answer correctness.}
\textbf{(a)} Across five Qwen2.5-7B GSM8K seed pairs, strict
training--request interactions vary across seeds, while full-response MV
interactions remain near zero.
\textbf{(b)} In content-matched SFT, changing only the final-answer marker
reverses convention preference across three seeds; the unchanged boxed-marker
control remains near baseline.
\textbf{(c)} Across three stratified human-calibration settings, strict-score
gains exceed committed-answer gains; intervals and identification ranges
reflect calibration uncertainty and unresolved-label sensitivity.
}
    \label{fig:main}
\end{figure*}

\paragraph{Behavioral and score interactions measure different outcomes.}
Across the full $1{,}319$-item set, payload interactions average $+47.08$
percentage points (seed SD $31.10$), versus $+42.70$ under strict scoring
(SD $31.38$) and $-0.20$ under MV (SD $0.89$).
Figure~\ref{fig:main}a visualizes the strict--MV contrast across seeds;
Table~\ref{tab:five-seed-interactions} reports the full summary.
Seed $85$ combines a small positive payload interaction ($+1.97$) with a
negative strict-score interaction ($-2.50$), showing that payload presence
and score need not move together.
These are descriptive summaries of five fixed seeds; small MV interactions
do not establish semantic equivalence.

\paragraph{Controlled interventions support learnable, familiarity-dependent reporting preferences.}
\label{sec:d1}\label{sec:manufactured}
Content-matched SFT tests preference acquisition directly:
$7{,}377$ paired SFT targets are byte-identical except for the final marker,
and the training prompts contain no format instruction.
Across three seeds, preference $100D$ shifts from $-55.90$ percentage points
initially to $+53.52$ in the \hash{} arm, while remaining $-55.40$ in the
boxed control.
The arm-to-arm contrast is $+108.92$ percentage points, with a cross-seed
range of $1.00$ point (Figure~\ref{fig:main}b).
Both arms reduce lenient accuracy from $69.95\%$ to approximately $55\%$,
showing that reporting preference can be acquired even while lenient answer
scores decline.
Table~\ref{tab:interventions} provides complementary reward and convention
controls.
Removing the \hash{} requirement while retaining answer correctness reduces
Qwen's strict gain from $+62.68$ to $+7.33$ percentage points across three
seeds ($88.3\%$ attenuation), and SmolLM2's from $+31.15$ to $-0.78$
points across two ($102.5\%$ attenuation).
A matched SmolLM2 seed-$83$ control similarly shifts payload interaction from
$+64.14$ points (marker-required) to $-6.52$ (answer-only; low initial
adherence; Appendix~\ref{app:smol-reward-crossed}).

\begin{table}[!htb]
\caption{\textbf{Reward and convention controls probe convention-sensitive gains.}
Gains are percentage points; $\rho$ is attenuation relative to control.}
\label{tab:interventions}
\centering
\footnotesize
\setlength{\tabcolsep}{4.0pt}
\renewcommand{\arraystretch}{1.06}

\begin{tabular*}{0.8\linewidth}{
@{\extracolsep{\fill}}
llrrrc
@{}}
\toprule

Intervention
& Model
& \multicolumn{2}{c}{Strict gain}
& $\rho$ (\%)
& Seeds \\
\cmidrule(lr){3-4}
&& Control & Treatment && \\

\midrule

\rowcolor{TableBand}
\multicolumn{2}{@{}l}{\textbf{A. Remove marker requirement}}
& \multicolumn{4}{r@{}}{\itshape reward control} \\

Answer only
& Qwen2.5 1.5B
& $+62.68$
& $+7.33$
& $88.3$
& $3$ \\

Answer only
& SmolLM2 1.7B
& $+31.15$
& $-0.78$
& $102.5$
& $2$ \\

\addlinespace[1pt]

\rowcolor{TableBand}
\multicolumn{2}{@{}l}{\textbf{B. Swap requested/rewarded convention}}
& \multicolumn{4}{r@{}}{\itshape convention control} \\

\hash{} $\rightarrow$ \boxedm{}
& Qwen2.5 1.5B$^\dagger$
& $+69.07$
& $+6.90$
& $90.0$
& $3$ \\

\hash{} $\rightarrow$ \boxedm{}
& SmolLM2 1.7B
& $+29.64$
& $+30.60$
& $-3.2$
& $2$ \\

\bottomrule
\end{tabular*}

\vspace{2pt}
\parbox{0.97\linewidth}{\scriptsize
Rows summarize treatment seeds; $^\dagger$ gives seed-$83$ Qwen point estimates;
the Seeds=$3$ entry includes two paired replications. Each arm uses validation matched to its own reward;
$\rho>100\%$ denotes treatment loss.
Additional controls appear in
Appendices~\ref{app:optimiser} and~\ref{app:horizon}.}
\end{table}

Changing both requested and rewarded conventions yields matched
Qwen2.5-1.5B gains of $+69.07$ percentage points under \hash{} and $+6.90$
under \boxedm{} despite similar validation endpoints.
The resulting $90.0\%$ attenuation replicates at $89.93\%$ and $90.00\%$
in two additional seeds.
SmolLM2 provides a contrasting case: gains are $+29.64$ and $+30.60$
percentage points when neither convention is initially well extracted.
Together, these controls support a familiarity-dependent interpretation rather
than a universal marker advantage.
Optimizer and longer-horizon controls retain roughly $85$--$90\%$ attenuation
(Appendices~\ref{app:optimiser} and~\ref{app:horizon}), while lenient gains
increase with training horizon, showing that score changes are not confined to
the strict reader.

\subsection{RQ2: Do strict-score gains imply better committed answers?}
\label{sec:results-obs}\label{sec:main}

\paragraph{Human calibration compares strict-score and committed-answer gains.}
\label{sec:human-calibration}
Calibration uses $625$ frozen responses from three settings, not the full
Qwen2.5-7B GSM8K crossed matrix. Two independent non-author annotators locked
commitments without reference answers, then judged correctness with the
references revealed; adjudication remained blinded to model condition and reader
scores, and no AI system supplied labels. Prespecified analyses retain \texttt{UNRESOLVED} cases via identification
ranges and simultaneous exact finite-population bounds, without post-label
resampling or exclusion (Appendix~\ref{app:human-calibration}).

\begin{table}[!htb]
\caption{\textbf{Strict-score gains exceed human-calibrated committed-answer gains.}
Gains and gaps are percentage points. ID denotes identification ranges over
unresolved labels; CIs are simultaneous $95\%$ finite-population bounds.}
\label{tab:human-semantic-calibration}

\centering
\small
\setlength{\tabcolsep}{3.8pt}
\renewcommand{\arraystretch}{1.08}

\begin{tabular*}{0.96\linewidth}{
@{\extracolsep{\fill}}
lccccc
@{}}
\toprule

&
\multirow{2}{*}{\shortstack{Strict gain\\$\Delta_S$}}
&
\multicolumn{2}{c}{Human gain $\Delta_H$}
&
\multicolumn{2}{c}{Strict--human gap $\Delta_S-\Delta_H$} \\

\cmidrule(lr){3-4}
\cmidrule(l){5-6}

Setting
&
&
ID range
&
$95\%$ CI
&
ID range
&
$95\%$ CI \\

\midrule

\rowcolor{TableBand}
\multicolumn{2}{@{}l}{\textbf{A. GSM8K}}
&
\multicolumn{4}{r@{}}{\itshape committed-answer calibration} \\

Qwen2.5 1.5B
& $+64.29$
& $+2.13$--$+4.84$
& {\color{TableCI}$[-27.22,+30.40]$}
& $+59.45$--$+62.16$
& $\mathbf{[+33.89,+91.51]}$ \\

SmolLM2 1.7B
& $+31.31$
& $+2.42$--$+9.80$
& {\color{TableCI}$[-20.05,+24.26]$}
& $+21.51$--$+28.89$
& $\mathbf{[+7.05,+51.36]}$ \\

\addlinespace[2pt]

\rowcolor{TableBand}
\multicolumn{2}{@{}l}{\textbf{B. MATH500}}
&
\multicolumn{4}{r@{}}{\itshape symbolic-transfer calibration} \\

Qwen2.5 7B
& $+8.53$
& $-9.74$--$-6.58$
& {\color{TableCI}$[-33.41,+6.27]$}
& $+15.11$--$+18.27$
& $\mathbf{[+2.26,+41.94]}$ \\

\bottomrule
\end{tabular*}

\vspace{2pt}
\parbox{0.97\linewidth}{\scriptsize
Strict gains use the full frozen response cache. MATH500 averages fixed
seeds $83$--$85$ against a shared initial policy. Bounds condition on frozen
caches and judgments, not retraining uncertainty. Correctness refers to
committed answers, not reasoning chains
(Appendix~\ref{app:human-calibration}).}
\end{table}

All three strict--human gain-gap intervals exclude zero, showing that strict-score
gains exceed corresponding committed-answer gains in the calibrated settings;
for Qwen2.5-1.5B, the lower bound exceeds half the observed strict gain.
Every human-gain interval nevertheless includes zero, so these data establish
neither positive committed-answer gains nor, for MATH500, a decline.
For GSM8K, both the lenient and MV gains fall within the corresponding human
identification ranges, but this neither validates those readers nor establishes unbiasedness.
Descriptive error-component and annotation-reliability diagnostics appear in
Tables~\ref{tab:human-reader-decomposition}
and~\ref{tab:human-annotation-reliability}.

\paragraph{Most strict-score gains coincide with closing the initial reader gap, while lenient and MV gains remain.}
\label{sec:validity}
Under matched evaluation, Qwen2.5-1.5B gains $+64.29$ percentage points under
strict scoring but only $+4.62$ under the frozen lenient reader; SmolLM2 gains
$+31.31$ and $+5.16$ points, respectively.
Net strict--lenient gap reduction accounts descriptively for $92.8\%$ and
$83.5\%$ of these gains, respectively, while the lenient-gain intervals
$[2.20,7.13]$ and $[2.65,7.58]$ percentage points exclude zero.

Fixed rephrasing and one-example prompting reduce Qwen's initial gap from
$59.67$ to $16.30$ and $13.19$ percentage points, with strict gains of
$+20.09$ and $+16.76$ points; greedy decoding preserves the qualitative
contrast (Appendix~\ref{app:matched}). Under the default prompt, Math-Verify
yields gains of $+3.87$ and $+4.09$ points for Qwen and SmolLM2, while OLMo MV
gains remain $+3.5$--$+4.4$ across three seeds (Table~\ref{tab:matched}); the
corresponding paired intervals exclude zero. Because prompt changes alter the
response distribution, these are interface-sensitivity checks rather than
semantic-equivalence claims. Full matched results appear in Appendix~\ref{app:matched}.

\subsection{RQ3: Does producing the requested marker preserve the reported value?}
\label{sec:transfer-tests}

Symbolic transfer tests whether producing the marker preserves the
value reported through it.

\paragraph{Transfer separates marker compliance from value preservation.}
\label{sec:math-transfer}
Under a hash request, the seed-$83$ boxed-trained checkpoint changes by
$-10.4$ percentage points under strict scoring
(95\% CI $[-14.6,-6.2]$), but only $+0.8$ under MV
($[-2.4,4.0]$).
Across the full crossed matrix, the interaction is $+9.4$ points under strict
scoring ($[3.6,15.4]$) but $-7.2$ under MV ($[-11.8,-2.6]$), so the near-zero
GSM8K MV interaction does not persist under MATH500 transfer. The multi-seed
contrast is shown in Figure~\ref{fig:math-transfer-gains} and
Appendix~\ref{app:qwen7-seeds}.

\paragraph{Human and response-level evidence supports the distinction.}
Human calibration gives a gain identification range of $[-9.74,-6.58]$
percentage points, but its simultaneous $95\%$ interval $[-33.41,+6.27]$
spans both signs (Table~\ref{tab:human-semantic-calibration}), so these data do
not establish an overall MATH500 decline. At the response level, $67$ of $500$
seed-$83$ hash-trained, hash-request responses contain a correct boxed value
but an incorrect hash value, including one deriving $\boxed{-50}$ before
emitting \hash{}~$50$. Thus, marker compliance need not preserve the reported
value (Appendix~\ref{app:math-transfer}).

\subsection{RQ4: Which interpretations survive boundary tests?}
\label{sec:boundary}

Table~\ref{tab:boundary-tests} summarizes the boundary checks that most directly
constrain the main interpretation.
\begin{table}[!htb]
\caption{\textbf{Boundary tests constrain the interpretation.}}
\label{tab:boundary-tests}

\centering
\footnotesize
\setlength{\tabcolsep}{4.0pt}
\renewcommand{\arraystretch}{1.04}

\begin{tabular*}{0.86\linewidth}{
@{\extracolsep{\fill}}
llrr
@{}}
\toprule

Test & Setting & \multicolumn{2}{c}{$\Delta$ (pp)} \\
\cmidrule(l){3-4}
&& Strict & Alt. reader \\
\midrule

\rowcolor{TableBand}
\multicolumn{4}{@{}l@{}}{\textbf{A. Reader dependence}\hfill\itshape MV readout} \\

Coverage ($k=256$)
& Qwen
& $+3.34$
& $-0.53$ \\

\multirow{2}{*}{Fresh arithmetic}
& Qwen
& $+87.03$
& $+0.39$ \\

& SmolLM2
& $+54.04$
& $+5.89$ \\

\addlinespace[1pt]
\rowcolor{TableBand}
\multicolumn{4}{@{}l@{}}{\textbf{B. Aligned-interface boundary}\hfill\itshape lenient readout} \\

Native MATH
& Qwen2.5 7B / 83--85
& $+2.2/+1.4/+0.0$
& $+2.2/+1.4/+0.4$ \\

\bottomrule
\end{tabular*}
\end{table}

OLMo-2 JSON/XML starts above $97\%$ adherence and yields only
$+0.38/+0.30/+0.68$-point payload interactions across three seeds
(Appendix~\ref{app:olmo-jsonxml}). Thus large interference is not inevitable
when the interface is aligned, while Table~\ref{tab:boundary-tests} shows that
reader-dependent gains can vanish.

\section{Related work}
\label{sec:related}

RLVR can correct prompt--format misalignment \citep{wu2025invisible}, while
verifier errors and reward hacking complicate score interpretation
\citep{huang2025accuracy}, and gains beyond format correction have also been
reported \citep{wang2026reinforcement}. Related studies examine robust verifiers
\citep{liu2025compassverifier}, SFT format imitation \citep{yang2025emperor},
instruction-following degradation \citep{fu2026scaling}, persistent training
conventions \citep{zhang2026inverse}, and prompt sensitivity
\citep{sclar2024quantifying,alzahrani2024benchmarks,mizrahi2024state}. We instead
cross rewarded and requested conventions, compare with the initial policy,
rescore the same outputs with fixed readers, and calibrate selected gains with
human judgments. Appendix~\ref{app:related} gives the detailed positioning.

\section{Conclusion}
\label{sec:conclusion}

Crossed evaluation shows that RLVR can reinforce reporting preference while
weakening request adherence: boxed training reduces requested hash-payload
presence in four of five Qwen2.5-7B GSM8K seeds, including under fixed
paraphrases. SFT, human calibration, and symbolic transfer distinguish learned
preference, committed-answer gains, marker compliance, and value preservation,
clarifying why interface choice belongs in the evaluation protocol. We therefore
recommend evaluating request adherence alongside convention-matched accuracy
and committed-answer correctness.

\clearpage
\subsubsection*{Reproducibility statement}
The appendix documents per-arm training and evaluation settings, reader
definitions, matched scores, uncertainty procedures, and the human-calibration
protocol. Retained supporting artifacts include frozen prompts, response-level
outputs and scores, sampling manifests and weights, source hashes, figure inputs,
and analysis code for reconstructing the reported analyses. Human calibration is documented in
Appendix~\ref{app:human-calibration}; relevant experimental and evaluation details
are provided throughout the supporting appendices.

\subsubsection*{Ethics statement}
This study uses publicly released models and benchmarks to improve the
interpretation of evaluation results. Human semantic labels concern the
answers in model-generated mathematical responses. No AI system produced item-level human labels, and the labels do not certify the reasoning chains.

\subsubsection*{Statement on LLM usage}
Generative AI assisted literature search, conceptual and experimental design,
implementation, synthetic arithmetic generation, analysis, manuscript revision,
and scientific graphics. AI-assisted outputs used in the analyses were independently
checked with the corresponding deterministic or human evaluation procedures;
no AI system produced or adjudicated the human-calibration labels.
The authors reviewed all AI-assisted work and take responsibility for the final
content and claims.


\appendix
\let\addcontentsline\AppendixAddContentsLine
\raggedbottom

\newpage
\appendix

\vspace{2em}
\begin{center}
    {\Large\textbf{Appendix}}
\end{center}
\vspace{2em}

\etocdepthtag.toc{appendix}
\etocsettagdepth{chapter}{none}
\etocsettagdepth{appendix}{subsection}
\tableofcontents

\section{Semantic calibration of committed answers}
\label{app:human-calibration}

\paragraph{Question and target population.}
This calibration asks whether reader-measured training gains agree with
changes in committed-answer correctness. It does not ask whether a correct
value appears anywhere in the response, whether the derivation is valid, or
whether latent reasoning ability has changed. Table~\ref{tab:human-semantic-calibration} reports the aggregate
calibration for the matched original-prompt GSM8K responses and the
Qwen2.5-7B hash-request MATH500 responses. Other prompts, OLMo, arithmetic,
coverage samples, and the boxed-trained 7B arm are outside this human
sample. Accordingly, the calibration does not cover the complete 7B crossed-request
matrix.

\paragraph{Frozen stratified sampling.}
The sample contains $625$ response observations from eight model states
(Table~\ref{tab:human-allocation}). Within each state, strata are the
correctness-bit combinations of strict/fallback/MV on GSM8K and strict/MV
on MATH500. Each nonempty stratum contributes a uniform sample without
replacement, capped at $30$ and $20$ responses, respectively. The $34$
nonempty strata include both agreement and disagreement states; all $36$
GSM8K fallback/MV disagreements are included. The sample was frozen before human outcomes were
observed, with no post-label resampling or exclusion. A pair of identical
response texts from one Smol initial/RL problem is retained as two distinct
checkpoint observations. There are no duplicate state/response identities.

\begin{table}[htbp]
\caption{\textbf{Human calibration sample and finite populations.}
Initial and trained states are sampled separately. The MATH500 initial
state is shared across all three seed contrasts rather than replicated
three times.}
\label{tab:human-allocation}
\centering
\footnotesize
\setlength{\tabcolsep}{7.5pt}
\renewcommand{\arraystretch}{1.05}
\begin{tabular*}{0.90\linewidth}{
@{\extracolsep{\fill}}
llrr
@{}}
\toprule
Dataset / model & State & Population & Sample \\
\midrule
GSM8K / Qwen2.5 1.5B
    & Initial      & $1319$ & $105$ \\
    & RL, seed $83$ & $1319$ & $61$ \\
GSM8K / SmolLM2 1.7B
    & Initial      & $1319$ & $106$ \\
    & RL, seed $83$ & $1319$ & $82$ \\
MATH500 / Qwen2.5 7B
    & Shared initial & $500$ & $69$ \\
    & RL, seed $83$ & $500$ & $62$ \\
    & RL, seed $84$ & $500$ & $70$ \\
    & RL, seed $85$ & $500$ & $70$ \\
\bottomrule
\end{tabular*}
\end{table}

\paragraph{Annotation execution and independence.}
\label{app:human-calibration-execution}
Annotator A was a doctoral researcher experienced in NLP and LLM
evaluation; annotator B was a graduate researcher trained in machine
learning and mathematical reasoning. A senior doctoral researcher
experienced in LLM evaluation adjudicated. All three were non-authors and were uninvolved in model training, checkpoint
selection, or reader development. No AI system produced, revised, or adjudicated item-level
human labels. No sampling, annotation, adjudication, or analysis rule changed after human
labels were inspected.

A and B independently completed two stages on separately ordered files.
Stage~1 showed only the anonymized sample ID, question, and complete
original response, withholding the reference answer, model/checkpoint
identity, reader outputs and the other annotator's labels.
Annotators recorded whether the response contained a clear committed
answer, no answer, an ambiguous commitment, or an unresolved conflict,
together with supporting text and self-correction/withdrawal flags. An
explicitly withdrawn answer is superseded by its correction; conflicting
unretracted answer fields are not resolved by selecting the
reference-matching one.

Both Stage~1 submissions were locked before Stage~2 revealed reference
answers and randomly permuted anonymous reader candidates; model identity and
reader scores remained hidden. Stage~2 assessed the locked commitment,
preserving signs, units, sets, ordered tuples, and multipart answers.
Suspected reference errors were retained as unresolved unless adjudicated.
Candidate fidelity was judged separately from agreement with the reference.

\paragraph{Adjudication and quality control.}
The adjudicator remained blinded to model condition and reader scores while
reviewing substantive A/B disagreements, flagged reference issues, and
explicit review requests. A fixed random review
covered $58$ initially concordant responses, approximately $10\%$ of that
pool; $2$ provisional consensus labels required revision. Ambiguous cases
remained \texttt{UNRESOLVED}. Original A/B labels were preserved for
pre-adjudication agreement analysis. Twenty-four separate synthetic
examples were used for protocol training and are excluded from the $625$
observations. 
\paragraph{Population weighting and unresolved commitments.}
For a fixed model state, let stratum $h$ have population size $N_h$ and
sample size $n_h$, and let $\mathcal S$ denote the sampled observations.
An observation $i$ in stratum $h(i)$ has inclusion probability
$n_{h(i)}/N_{h(i)}$ and prespecified weight
$w_i=N_{h(i)}/n_{h(i)}$. With $N=\sum_h N_h$, define $z_i^L=1$ for a
definitely correct committed answer and $0$ otherwise. Define $z_i^U=1$
for either a definitely correct or final \texttt{UNRESOLVED} answer, and
$0$ for a definitely incorrect answer or no answer. Estimated
committed-answer accuracy is therefore partially identified by
\begin{equation}
[\widehat H^L,\widehat H^U]
=
\left[
\frac{1}{N}\sum_{i\in\mathcal S} w_i z_i^L,\;
\frac{1}{N}\sum_{i\in\mathcal S} w_i z_i^U
\right].
\label{eq:human-identification}
\end{equation}
Unresolved cases are neither dropped nor silently marked incorrect. The
estimated human-gain bounds are
\[
\left[
\widehat H_{\rm RL}^{L}-\widehat H_{0}^{U},\;
\widehat H_{\rm RL}^{U}-\widehat H_{0}^{L}
\right].
\]
For MATH500, the fixed-seed contrast is
$(H_{83}+H_{84}+H_{85})/3-H_0$, with the shared initial estimate entering
once rather than being replicated across seed contrasts. Reader gains use
the complete frozen response populations. For a fixed reader gain
$\Delta_E$, subtracting the human-gain bounds reverses their order,
yielding
$[\Delta_E-\widehat{\Delta}_H^U,\,
  \Delta_E-\widehat{\Delta}_H^L]$
for the reader--human gain gap.

\paragraph{Confidence bounds and scope.}
The prespecified analysis inverts the hypergeometric distribution within
each stratum for the declared binary endpoints, including lower/upper
human correctness, unresolved status, agreement, and candidate fidelity.
For $M$ noncensus endpoint-by-stratum cells, the per-cell error budget is
$0.05/M$; census cells have no sampling uncertainty. A union bound then
gives simultaneous coverage, which is propagated through
population-weighted sums and signed contrasts.

In Table~\ref{tab:human-semantic-calibration}, the ID-range columns report
unresolved-label partial identification at the weighted-estimate level,
whereas the simultaneous $95\%$ confidence bounds additionally incorporate
finite-population sampling uncertainty. They are not confidence intervals around an imputed midpoint. These conservative bounds condition
on the frozen response caches and human judgments. They do not cover new
questions, fresh generations, retraining variability, or systematic errors
shared by the annotators.

\paragraph{What the calibration establishes.}
All three strict--human gain-gap simultaneous $95\%$ confidence intervals
exclude zero on the positive side, whereas every human-gain confidence
interval includes zero. The supported conclusion is a discrepancy between strict-score progress and
committed-answer progress, not a precise estimate of remaining semantic
improvement or a confirmed MATH500 decline. The smaller GSM8K lenient-reader gains and the MATH500
mean MV decline are compatible with the corresponding human identification
ranges, but such compatibility is weaker than validating either reader on
individual responses.

Equation~\ref{eq:readerbias} expresses the reader--human gain gap as a
change in net reader bias relative to committed-answer correctness.
Table~\ref{tab:human-reader-decomposition} reports the corresponding
strict-reader components: reduction in missed correct credit and increase
in false credit. Their identification endpoints reproduce the gain-gap
endpoints in all three calibrated settings. The missed-credit reduction is
larger throughout the reported estimate ranges, while positive estimated
false-credit increases remain for Smol and MATH500. This pattern is descriptive, not evidence of componentwise significance or
population-level dominance; component confidence bounds are not reported.
A negative false-credit change does not imply that the trained reader has
no false positives. Neither component is a causal mediation effect of
reporting behavior, and the decomposition does not certify the alternative readers'
fidelity or reasoning-chain validity.

\paragraph{Agreement and unresolved uncertainty.}
Table~\ref{tab:human-annotation-reliability} reports original independent
A/B category agreement and final unresolved mass. Correctness agreement
ranges from $92.22\%$ to $96.80\%$, whereas answer-status agreement is
$100\%$ in every state. The latter concerns status categories such as
clear or ambiguous commitment, not agreement on the numerical answer.
Final unresolved mass is a distinct quantity: both annotators can agree
that a response is unresolved, while adjudication can either resolve a
disagreement or retain uncertainty. It therefore need not equal or be bounded by the pre-adjudication disagreement
rate. The $2$ revisions
among $58$ randomly reviewed concordant responses are unweighted review counts, not
an estimated population error rate. Neither agreement nor this review
excludes systematic shared error.

The unresolved masses also reproduce the widths of the estimated
human-gain ranges. If
$U_m=\widehat H_m^U-\widehat H_m^L$, the gain-range width is
$U_{\rm RL}+U_0$. The two GSM8K widths are
$2.71+0.00=2.71$ and $3.42+3.96=7.38$ points. For the MATH500 fixed-seed
mean, the width is
$(5.52+3.09+0.88)/3+0.00\simeq3.16$ points, with one shared initial.
These reproduce Table~\ref{tab:human-semantic-calibration} to rounding.
They quantify unresolved-label sensitivity rather than sampling
uncertainty; a reported $0.00\%$ unresolved mass does not establish zero
population uncertainty.

\paragraph{Reporting scope and reproducibility.}
These diagnostics are descriptive and do not alter the frozen confidence
family. We retain the frozen sample manifest, original
independent A/B labels, adjudication and quality-control records, final item-level
labels, sampling weights, and analysis code for reconstruction by
\texttt{sample\_id}. The retained A/B labels support the pre-adjudication
agreement estimates; final decisions support the gain and error estimates.

\begin{table}[!htbp]
\caption{\textbf{Estimated error components of the strict--human gain gap.}
Population-weighted identification ranges over unresolved labels, in
percentage points; these are estimates, not confidence intervals.
The two signed components account for the gap in
Table~\ref{tab:human-semantic-calibration}.}
\label{tab:human-reader-decomposition}
\label{tab:human-reader-diagnostics}
\centering\small
\setlength{\tabcolsep}{4.5pt}
\renewcommand{\arraystretch}{1.12}
\begin{tabular*}{\linewidth}{@{\extracolsep{\fill}}lccc@{}}
\toprule
Setting & \shortstack{Missed-correct reduction\\$\mathrm{FN}_0-\mathrm{FN}_{\rm RL}$}
 & \shortstack{False-credit increase\\$\mathrm{FP}_{\rm RL}-\mathrm{FP}_0$}
 & \shortstack{Strict--human gap\\$\Delta_S-\Delta_H$}\\
\midrule
GSM8K / Qwen2.5 1.5B & $[+59.82,+62.16]$ & $[-0.37,+0.00]$ & $[+59.45,+62.16]$\\
GSM8K / SmolLM2 1.7B & $[+18.76,+26.14]$ & $+2.75$ & $[+21.51,+28.89]$\\
MATH500 / Qwen2.5 7B & $[+9.90,+12.77]$ & $[+5.21,+5.50]$ & $[+15.11,+18.27]$\\
\bottomrule
\end{tabular*}
\par\vspace{3pt}
\begin{minipage}{\linewidth}\scriptsize
FP and FN are joint error probabilities over all responses, not conditional
FPR/FNR. Either change may be negative. MATH500 averages fixed seeds
$83$--$85$ against one shared initial. The identity is descriptive error
accounting, not a causal mediation analysis. Componentwise confidence
bounds are not reported here; the table supports no componentwise
significance or population-dominance claim.
\end{minipage}
\end{table}

\begin{table}[!htbp]
\caption{\textbf{Independent annotation agreement and residual uncertainty.}
Population-weighted descriptive estimates under the frozen stratified
design. Correctness agreement uses the original A/B labels before
adjudication; final unresolved mass is measured after third review.
Confidence intervals are not reported in this display.}
\label{tab:human-annotation-reliability}
\label{tab:human-annotation-quality}
\centering\small
\setlength{\tabcolsep}{6pt}
\renewcommand{\arraystretch}{1.10}
\begin{tabular*}{\linewidth}{@{\extracolsep{\fill}}llcc@{}}
\toprule
Setting & State & \shortstack{Correctness\\agreement (\%)}
 & \shortstack{Final unresolved\\mass (\%)}\\
\midrule
GSM8K / Qwen2.5 1.5B & Initial & $96.80$ & $2.71$\\
 & RL / $83$ & $95.85$ & $0.00$\\
\addlinespace[2pt]
GSM8K / SmolLM2 1.7B & Initial & $96.37$ & $3.42$\\
 & RL / $83$ & $94.74$ & $3.96$\\
\addlinespace[2pt]
MATH500 / Qwen2.5 7B & Shared initial & $96.54$ & $0.00$\\
 & RL / $83$ & $95.78$ & $5.52$\\
 & RL / $84$ & $93.61$ & $3.09$\\
 & RL / $85$ & $92.22$ & $0.88$\\
\bottomrule
\end{tabular*}
\par\vspace{3pt}
\begin{minipage}{\linewidth}\scriptsize
Correctness agreement is exact A/B category agreement, including unresolved
labels, over all sampled responses with population weights.
Answer-status agreement is $100\%$ in every state; agreement on a status
category does not imply agreement on the answer value or its correctness.
The blinded adjudicator rechecked $58$ initially concordant responses
(approximately $10\%$), revising $2$ provisional consensus labels.
These are unweighted review counts, not a population error rate. Unresolved
mass is not the disagreement rate; agreement does not exclude shared error.
\end{minipage}
\end{table}

\clearpage
\section{Convention learning, request dependence, and transfer}
\label{app:convention-transfer}

\subsection{Qwen2.5-7B training-seed replication}
\label{app:qwen7-seeds}

\paragraph{Scope and training controls.}
We add full-parameter Qwen2.5-7B-Instruct hash-reward runs at seeds $84/85$
to the completed seed-$83$ run. Each run completes $100$
updates on four GPUs. We retain one final checkpoint per run, comprising
four FSDP shards and a verified merged model; the same four GPUs are reused
serially for the new runs. No checkpoint is selected by evaluation
accuracy. Configurations match seed $83$ except for four fields: data seed,
rollout seed, experiment name, and output directory. Initial weights, the
$7{,}377$ frozen GSM8K training rows, $1{,}319$ validation rows, reward
code, fixed nonuniform prompt proposal, optimizer, and all other settings
are held fixed; hashes and training-completion records are retained. The
full recipe is in Appendix~\ref{app:train-config}.

This replication initially repeats hash-reward training only. The later GSM8K
comparison adds boxed-trained seeds $84/85$
(Appendix~\ref{app:qwen7-crossed-seeds}); the full MATH500 crossed-training
panel and reward-program re-scoring remain seed-$83$ comparisons.

\paragraph{Reused evidence and frozen evaluation.}
The replication was motivated by the seed-$83$ results. Its training
manifest and scope were fixed before the additional training runs. The final
evaluation-source manifest was frozen during seed-$84$ training and before
any evaluation of the new runs, retaining the existing prompts, items,
readers, and decoding settings. Initial and seed-$83$ responses are reused
byte-for-byte; they are neither regenerated nor counted as new evidence.

The retained analysis contains $16$ cells and $14{,}552$ responses:
$4{,}000$ MATH500 and $10{,}552$ GSM8K responses, split equally between
reused and newly generated evidence ($7{,}276$ each). The boxed-trained
model's earlier crossed-panel cells are outside this count. The same three
hash-trained checkpoints serve both datasets; they are not six independent
training runs. No new prompt search, greedy panel, or multi-sample coverage
experiment is part of this replication.

MATH500 reuses its original generation worker and scorer with the canonical
initial tokenizer. GSM8K retains its model-local tokenizer path, with exact
encoded-token parity verified on all $2{,}638$ hash/boxed prompts for both
new checkpoints. MATH500 uses output/context limits of
$8{,}192/12{,}288$, batches of $32$, and a maximum of $64$ sequences;
GSM8K uses $640/4{,}096$, batches of $16$, and a maximum of $256$
sequences. All evaluations use bfloat16, $T=0.6$, top-$p=0.95$, one
response per item, and GPU memory utilization $0.8$. Item seeds are
$20260909+i$ for MATH500 and $20260908+i$ for GSM8K, with no input
truncation. Runtime versions are
vLLM 0.11.0 \citep{kwon2023efficient}, Transformers 4.57.6, PyTorch 2.8.0+cu128,
Math-Verify 0.9.0, SymPy 1.14.0,
\texttt{latex2sympy2\_extended} 1.11.0, and NumPy 1.26.4.

\begin{table}[htbp]
\caption{\textbf{Complete secondary evaluation of the hash-trained checkpoints.}
Exact correct counts for requested-format strict ($S$) and full-response MV.
Truncation counts are hash/boxed. The initial comparator is shared across
training seeds. The boxed-trained intervention arm is outside this panel.}
\label{tab:qwen7-seed-secondary}\centering\footnotesize\setlength{\tabcolsep}{11pt}
\begin{tabular}{@{}lrrrrr@{}}\toprule
 & \multicolumn{2}{c}{Hash request} & \multicolumn{2}{c}{Boxed request} & \\
\cmidrule(lr){2-3}\cmidrule(lr){4-5}
Checkpoint & $S$ & MV & $S$ & MV & Trunc. (H/B)\\\midrule
\rowcolor{TableBand}\multicolumn{6}{@{}l}{\textbf{GSM8K ($n=1319$ per cell)}}\\
Initial & 1025 & 1208 & 1209 & 1213 & 2/3\\
Seed 83 & 1208 & 1208 & 1217 & 1219 & 5/6\\
Seed 84 & 1213 & 1210 & 1171 & 1211 & 2/3\\
Seed 85 & 1218 & 1217 & 1215 & 1216 & 4/6\\
\rowcolor{TableBand}\multicolumn{6}{@{}l}{\textbf{MATH500 ($n=500$ per cell)}}\\
Initial & 265 & 367 & 375 & 377 & 3/2\\
Seed 83 & 259 & 333 & 383 & 384 & 5/2\\
Seed 84 & 320 & 312 & 367 & 369 & 0/0\\
Seed 85 & 344 & 326 & 380 & 380 & 3/1\\
\bottomrule\end{tabular}\end{table}

\paragraph{Secondary endpoints and adverse outcomes.}
Table~\ref{tab:qwen7-seed-secondary} reports all secondary cells; paired
hash-request gains appear in Table~\ref{tab:qwen7-seed-gains}. GSM8K
strict scoring reads the first numeric hash answer. Boxed strict scoring
extracts the first balanced box and then its last numeric token, with
tolerance $10^{-3}$; neither reader reproduces the last-$300$-character,
exact-string training reward. MATH500 uses its frozen symbolic payload
readers, while full-response MV is scored separately on every saved
response. These readers are neither nested nor semantic ground truth.

Large GSM8K hash-strict gains coexist with small MV changes in every
training seed; all three MV intervals include zero, which does not
establish equivalence. The pattern also does not imply improvement under
every request: under the GSM8K boxed request, strict accuracy falls
$2.88$ points in seed $84$ ($95\%$ CI $[-4.40,-1.36]$), while MV
changes by $-0.15$ points ($[-1.36,1.06]$).

\begin{table}[htbp]
\caption{\textbf{Hash-request gains remain strongly reader-dependent.}
Trained-minus-initial changes in percentage points, with $95\%$ paired-item
intervals ($20000$ resamples, seed $20260910$). All are secondary endpoints
conditional on fixed checkpoints and responses.}
\label{tab:qwen7-seed-gains}\centering\footnotesize\setlength{\tabcolsep}{9pt}
\begin{tabular}{@{}llcc@{}}\toprule
Dataset & Training seed & Strict gain & MV gain\\\midrule
GSM8K & 83 & $+13.87\;[11.68,16.07]$ & $+0.00\;[-1.21,1.21]$\\
 & 84 & $+14.25\;[12.05,16.45]$ & $+0.15\;[-1.14,1.44]$\\
 & 85 & $+14.63\;[12.51,16.83]$ & $+0.68\;[-0.53,1.90]$\\
\midrule
MATH500 & 83 & $-1.20\;[-6.20,4.00]$ & $-6.80\;[-10.60,-3.00]$\\
 & 84 & $+11.00\;[6.40,15.60]$ & $-11.00\;[-15.20,-7.00]$\\
 & 85 & $+15.80\;[11.40,20.40]$ & $-8.20\;[-11.80,-4.80]$\\
\bottomrule\end{tabular}\end{table}

Under MATH500 hash requests, MV falls in every seed, whereas strict
accuracy increases in both new seeds. Native-boxed MV changes are
$+1.4/-1.6/+0.6$ points for seeds $83/84/85$. This reproduces reader- and request-sensitivity under the fixed protocol, not
a shared latent mechanism or a causal estimate of reasoning change.

All secondary comparisons retain every item and use $20{,}000$ shared
item-bootstrap resamples with seed $20260910$, under the same conditional fixed-checkpoint interpretation used throughout the transfer analysis.
The original crossed MATH500 analysis retains its earlier bootstrap seed
$20260909$; minor differences in percentile endpoints reflect resampling,
not changed responses or scores. Truncation totals are $16/4{,}000$ for
the four MATH500 checkpoint states and $31/10{,}552$ for GSM8K. The
retained evidence includes input, raw-response, and score manifests; token
counts; termination reasons; configuration checks; and distinct
model-weight hashes. Independent local re-analysis reproduces all reported
cell counts and contrasts. None of these checks supplies human semantic
labels.

\subsection{Paired hash--boxed replication of the GSM8K crossed comparison}
\label{app:qwen7-crossed-seeds}

\paragraph{Scope and matching.}
The complete GSM8K comparison pairs hash- and boxed-trained policies
within five seeds, $83$--$87$. After the original three pairs, both arms
of seeds $86$ and $87$ were specified together and run to completion; all
earlier outcomes, including seed $85$, remain in the analysis. Each run uses full-parameter Qwen2.5-7B-Instruct training with the same initial
weights, $7{,}377$ frozen training items, and $100$ updates. Within each pair,
configurations differ only in convention-specific data, reward and output
identifiers. No checkpoint or seed is selected by evaluation accuracy.
The eight extension cells add $10{,}552$ responses to $15{,}828$ original
trained responses. Every cell uses the same $1{,}319$ GSM8K items,
$T=0.6$, top-$p=0.95$, a $640$-token output limit, and frozen readers.
This extension adds no MATH500 generations or human semantic labels.

\paragraph{Paired contrast and reporting scope.}
For reader $R$ and seed $s$, the interaction in percentage points is
\[
I_{R,s}=100\big[(a^{R,s}_{H,h}-a^{R,s}_{B,h})
                    -(a^{R,s}_{H,b}-a^{R,s}_{B,b})\big],
\]
where $a$ is accuracy, $H/B$ denote training conventions and $h/b$ denote requests.
Table~\ref{tab:five-seed-interactions} and Figure~\ref{fig:main}a report
every fixed seed. Means weight the five seeds equally; sample SD uses
denominator $5-1$. These descriptive summaries do not support inference over a training-seed
population. The full-sample reporting analysis below uses the same frozen detector
and per-response records.

\begin{table}[htbp]
\centering\footnotesize
\caption{\textbf{Complete five-seed GSM8K interactions.} All $20$ trained
cells contain $1319$ responses. $P$ measures requested payload presence,
$q$ frozen-reader extractability, $S$ strict accuracy, and MV full-response
Math-Verify. All entries are percentage points. Mean and sample SD
describe the five fixed seeds; no outcome is excluded.}
\label{tab:five-seed-interactions}
\begin{tabular*}{0.88\linewidth}{@{\extracolsep{\fill}}llrrrr@{}}
\toprule Seed & Cohort & $I_P$ & $I_q$ & $I_S$ & $I_{\rm MV}$\\
\midrule
83 & Original & $+39.80$ & $+39.88$ & $+35.18$ & $-0.99$\\
84 & Original & $+74.30$ & $+77.63$ & $+71.80$ & $+0.23$\\
85 & Original & $+1.97$ & $-2.50$ & $-2.50$ & $-0.45$\\
86 & Extension & $+40.64$ & $+40.86$ & $+35.63$ & $-0.91$\\
87 & Extension & $+78.70$ & $+79.00$ & $+73.39$ & $+1.14$\\
\midrule
\multicolumn{2}{l}{Mean} & $+47.08$ & $+46.97$ & $+42.70$ & $-0.20$\\
\multicolumn{2}{l}{Sample SD} & $31.10$ & $33.55$ & $31.38$ & $0.89$\\
\bottomrule\end{tabular*}\end{table}

\paragraph{Seed-dependent magnitude and direction.}
For original seeds $83/84/85$, the strict interactions are
$+35.18/+71.80/-2.50$ pp. Added seeds $86/87$
give $+35.63/+73.39$ pp. Thus four tested seeds show large positive interactions; seed $85$ instead
has a small negative interaction, with conditional item interval
$[-4.55,-0.53]$.
The reporting analysis below shows substantial request adaptation in
seed $85$. Thus the conflict is a possible outcome under this recipe,
with substantial training-seed variation.

The strict intervals use $20{,}000$ paired-item bootstrap draws
(RNG seed $20260913$), sharing sampled item IDs across cells and seeds.
They condition on fixed checkpoints and responses and do not quantify
retraining uncertainty. 
\paragraph{MV results and analysis retained.}
The original MV interactions are $-0.99$, $+0.23$ and $-0.45$ pp for
seeds $83/84/85$. Their paired-item $95\%$ intervals all include zero
in the retained original analysis. The original mean (sample seed SD) is
$-0.40$ ($0.61$) pp, with conditional item interval $[-1.34,0.53]$.
The interactions remain near zero with no consistent direction. Small MV interactions do not establish equal scores
or semantic ability: in seed $85$, hash training exceeds boxed training
by $1.21/1.67$ MV points under hash/boxed requests. MV is an operational
reader, not human ground truth. Human calibration does not cover the
complete crossed-request matrix.

The retained MV intervals use $20{,}000$ paired-item bootstrap resamples
(RNG seed $20260913$), sharing sampled item IDs across all cells and seeds.
They condition on the fixed model pairs and do not quantify retraining
uncertainty. 
\paragraph{Full-sample reporting analysis.}
\label{app:reporting-payload}
We analyze all $26{,}380$ trained responses ($5$ seeds $\times$ $2$ training
conventions $\times$ $2$ requests $\times$ $1{,}319$ items), together with $2{,}638$ initial-checkpoint responses ($2$ requests $\times$ $1{,}319$ items), without filtering
on correctness, consulting reference answers, or generating new outputs.
The detector was frozen before the five-seed analysis and then applied to the
extension; truncated responses are included. Each record contains
only the assistant completion, rather than the input prompt.

We measure two distinct properties. Frozen-reader extractability $q_c$
records whether the requested-convention reader returns a numeric value,
before comparing it with a reference. The hash reader takes the first
numeric match after \hash{}; the boxed reader takes the last numeric token
inside the first balanced box, retaining the original reader definitions.
Separately, the payload analysis scans every occurrence: an exact four-hash
marker must be followed on the same line by a numeric prefix, allowing
outer currency, math, or Markdown delimiters; a boxed payload must have
balanced braces and nonempty content. This rule excludes nonnumeric hash
headings, quoted hash-request mentions, and empty boxed templates.
All nonempty balanced boxes in these responses contain numeric content.
One qualifying occurrence suffices, regardless of answer correctness.

Table~\ref{tab:reporting-payload-counts} partitions responses by these
payload indicators. ``Both'' contributes to requested-payload presence:
neither prompt forbids the other convention, and a box can appear within
the derivation. The analysis measures syntactic reporting, not semantic
commitment or agreement between payloads. Presence can exceed $q_c$ when
an earlier empty box obstructs the frozen first-box reader even though a
later box carries content. The distinction is visible in seed $85$ under
boxed requests; it prevents conflating reader failure with marker omission.

In the new boxed-trained seeds, boxed-only responses account for
$530/534$ (seed $86$) and $1038/1041$ (seed $87$) hash-payload omissions,
consistent with $523/526$ and $980/981$ in the original deteriorating
seeds. Truncation counts in the eight extension cells are
$1/5/4/2$ for seed $86$ and $1/1/3/4$ for seed $87$, ordered as
hash training/hash request, hash/boxed, boxed/hash, boxed/boxed.
No truncated response is excluded.

\begin{table}[htbp]
\caption{\textbf{Full-sample reporting categories, independent of correctness.}
Each row contains $1319$ responses. Four mutually exclusive payload
categories sum to $1319$; $q_c$ is the number extractable by the frozen
requested-convention reader. Requested-payload presence is requested-only
plus both.}
\label{tab:reporting-payload-counts}
\centering\footnotesize
\setlength{\tabcolsep}{3pt}
\begin{tabular*}{0.97\linewidth}{@{\extracolsep{\fill}}lllrrrrr@{}}
\toprule
Seed & Training & Request & Requested only & Other only & Both & Neither & $q_c$\\
\midrule
--- & Initial & Hash & 1197 & 31 & 62 & 29 & 1109\\
--- & Initial & Boxed & 1316 & 0 & 0 & 3 & 1311\\
\midrule
83 & Hash & Hash & 1009 & 1 & 306 & 3 & 1315\\
83 & Hash & Boxed & 1314 & 0 & 0 & 5 & 1312\\
83 & Boxed & Hash & 75 & 523 & 718 & 3 & 792\\
83 & Boxed & Boxed & 1317 & 0 & 0 & 2 & 1315\\
\midrule
84 & Hash & Hash & 1317 & 0 & 0 & 2 & 1317\\
84 & Hash & Boxed & 1316 & 0 & 0 & 3 & 1271\\
84 & Boxed & Hash & 101 & 980 & 237 & 1 & 338\\
84 & Boxed & Boxed & 1317 & 0 & 0 & 2 & 1316\\
\midrule
85 & Hash & Hash & 1315 & 0 & 0 & 4 & 1315\\
85 & Hash & Boxed & 1313 & 0 & 0 & 6 & 1312\\
85 & Boxed & Hash & 1244 & 26 & 48 & 1 & 1292\\
85 & Boxed & Boxed & 1316 & 0 & 0 & 3 & 1256\\
\midrule
86 & Hash & Hash & 1318 & 0 & 0 & 1 & 1318\\
86 & Hash & Boxed & 1314 & 0 & 0 & 5 & 1312\\
86 & Boxed & Hash & 367 & 530 & 418 & 4 & 784\\
86 & Boxed & Boxed & 1317 & 0 & 0 & 2 & 1317\\
\midrule
87 & Hash & Hash & 1202 & 0 & 116 & 1 & 1318\\
87 & Hash & Boxed & 1318 & 0 & 0 & 1 & 1313\\
87 & Boxed & Hash & 48 & 1038 & 230 & 3 & 277\\
87 & Boxed & Boxed & 1316 & 0 & 0 & 3 & 1314\\
\bottomrule
\end{tabular*}
\end{table}

\paragraph{Initial-to-trained comparison.}
\label{app:reporting-baseline}
The shared initial checkpoint supplies one cached response per item and
request, counted once across the five training-seed comparisons. Input
content hashes, generation settings, item IDs, and per-item input token
counts match the trained cells. We apply the same payload detector and
frozen readers, including truncated responses. Table~\ref{tab:reporting-baseline}
reports the initial rates, trained rates, and same-item changes.
``Lost'' means the initial response has the requested payload and the
trained response does not; ``gained'' is the reverse. The net change in
percentage points is $100(\mathrm{gained}-\mathrm{lost})/1{,}319$. These compare saved responses on the same problems, not trajectories of a
single response or estimates of retraining uncertainty.

For hash requests, initial payload presence is $1{,}259/1{,}319$ ($95.45\%$),
versus frozen extractability $1{,}109/1{,}319$ ($84.08\%$). The $150$ differing
responses all begin the hash payload with a currency prefix, accepted
by the payload rule but missed by the frozen numeric reader. Thus this
initial extraction gap is not counted as missing requested payloads.
For boxed requests, initial payload presence is $1{,}316/1{,}319$ ($99.77\%$).
Boxed training reduces hash-payload presence by $35.33/69.83/35.94/74.37$
points in seeds $83/84/86/87$ but increases it by $2.50$ points in seed $85$.
The five-seed mean change is $-42.59$ pp. The large failures therefore reflect deterioration from initial request
adherence rather than persistence of the initial payload-omission rate.

For a correctness-independent crossed contrast, substitute requested-payload
frequency $P_{t,r}$ for accuracy in Equation~\ref{eq:crossinteraction}:
$I_P=100[(P_{H,h}-P_{B,h})-(P_{H,b}-P_{B,b})]$ in percentage points.
This gives $+39.80$, $+74.30$, $+1.97$, $+40.64$, and $+78.70$ pp
in seeds $83$--$87$. These descriptive behavioral
contrasts differ from strict-score interactions and do not measure the
correctness or consistency of values carried by the payloads.

\begin{table}[htbp]
\caption{\textbf{Requested reporting before and after training.}
Each cell has $1319$ saved responses. Payload and $q_c$ are percentages;
$q_c$ retains the frozen requested-convention reader. Lost/gained count
same-item payload changes from the shared initial cache; $\Delta P$ is
the net percentage-point change. The initial rows are shared baselines,
not separate replications.}
\label{tab:reporting-baseline}
\centering\footnotesize
\setlength{\tabcolsep}{4pt}
\begin{tabular*}{0.97\linewidth}{@{\extracolsep{\fill}}lllrrrrr@{}}
\toprule
Seed & Training & Request & Payload (\%) & $q_c$ (\%) & Lost & Gained & $\Delta P$ (pp)\\
\midrule
--- & Initial & Hash & 95.45 & 84.08 & --- & --- & ---\\
--- & Initial & Boxed & 99.77 & 99.39 & --- & --- & ---\\
\midrule
83 & Hash & Hash & 99.70 & 99.70 & 3 & 59 & $+4.25$\\
83 & Hash & Boxed & 99.62 & 99.47 & 4 & 2 & $-0.15$\\
83 & Boxed & Hash & 60.12 & 60.05 & 495 & 29 & $-35.33$\\
83 & Boxed & Boxed & 99.85 & 99.70 & 1 & 2 & $+0.08$\\
\midrule
84 & Hash & Hash & 99.85 & 99.85 & 1 & 59 & $+4.40$\\
84 & Hash & Boxed & 99.77 & 96.36 & 2 & 2 & $+0.00$\\
84 & Boxed & Hash & 25.63 & 25.63 & 932 & 11 & $-69.83$\\
84 & Boxed & Boxed & 99.85 & 99.77 & 1 & 2 & $+0.08$\\
\midrule
85 & Hash & Hash & 99.70 & 99.70 & 3 & 59 & $+4.25$\\
85 & Hash & Boxed & 99.55 & 99.47 & 4 & 1 & $-0.23$\\
85 & Boxed & Hash & 97.95 & 97.95 & 27 & 60 & $+2.50$\\
85 & Boxed & Boxed & 99.77 & 95.22 & 3 & 3 & $+0.00$\\
\midrule
86 & Hash & Hash & 99.92 & 99.92 & 0 & 59 & $+4.47$\\
86 & Hash & Boxed & 99.62 & 99.47 & 3 & 1 & $-0.15$\\
86 & Boxed & Hash & 59.51 & 59.44 & 503 & 29 & $-35.94$\\
86 & Boxed & Boxed & 99.85 & 99.85 & 2 & 3 & $+0.08$\\
\midrule
87 & Hash & Hash & 99.92 & 99.92 & 1 & 60 & $+4.47$\\
87 & Hash & Boxed & 99.92 & 99.55 & 1 & 3 & $+0.15$\\
87 & Boxed & Hash & 21.08 & 21.00 & 994 & 13 & $-74.37$\\
87 & Boxed & Boxed & 99.77 & 99.62 & 3 & 3 & $+0.00$\\
\bottomrule
\end{tabular*}
\end{table}

\subsection{Fixed wording check across five seed pairs}
\label{app:wording-check}

\paragraph{Fixed sample, requests, and checkpoints.}
Before generation, we fixed a uniform sample of $300$ GSM8K test IDs (sampling seed $20260919$), the original request and two
near-equivalent paraphrases, and all eleven checkpoints: one initial
model and both training conventions for seeds $83$--$87$. The question,
chat template and system message are held fixed; only the final request
suffix changes. The original suffix asks, ``Let's think step by step
and output the final answer after `\hash{}'.'' Paraphrase A asks,
``Think through the problem step by step. Give the final answer after
`\hash{}'.'' Paraphrase B asks, ``Work out the solution step by step
and present the final answer after `\hash{}'.'' Each boxed counterpart
replaces ``after `\hash{}''' with ``within \boxedm{}.''

Each cell has one response per item, $T=0.6$, top-$p=0.95$, a $640$-token
generation limit, and a $4096$-token context. Per-item generation
seeds follow $20260908+\text{original item ID}$. The original-template
responses are subsets of the frozen full-test cache, not regenerated;
the two paraphrases add $13200$ responses to $6600$ reused responses,
for $66$ cells and $19800$ responses. No item, seed, template or
checkpoint was replaced after inspecting results. The common initial
cache is shared across training seeds and counted once per template.

\paragraph{Within-template adaptation and crossed contrasts.}
All cells use the same frozen payload detector and readers as the
full-test analysis. For each template we subtract its own initial-model
rate: the initial hash-payload rates are $96.00\%$, $86.67\%$, and
$84.33\%$ for original, A, and B. This controls descriptively for each wording's different baseline adherence.
Table~\ref{tab:wording-seed-contrasts} reports every seed/template result;
Table~\ref{tab:wording-summary} summarizes the five seeds separately
within each template. The same four boxed-trained seeds deteriorate
under all templates, whereas seed $85$ improves under all three.
Seed $85$'s strict interaction changes sign across templates, showing
that this score contrast and its behavioral adaptation are distinct.

\begin{table}[htbp]\centering\footnotesize
\caption{\textbf{Fixed wording summaries across all five seeds.}
Entries are mean (sample seed SD), in percentage points. $\Delta P_{B,h}$
is boxed-trained minus initial hash-payload presence within each template.
$I$ denotes the training-by-request interaction. Each template uses the
same $300$ items; the original row is not the full $1319$-item result.}
\label{tab:wording-summary}
\begin{tabular*}{\linewidth}{@{\extracolsep{\fill}}lrrrr@{}}
\toprule Template & $\Delta P_{B,h}$ & $I_P$ & $I_S$ & $I_{\rm MV}$\\\midrule
Original & $-41.53$ (30.73) & $+45.67$ (30.69) & $+39.87$ (31.22) & $-0.27$ (1.38)\\
A & $-34.27$ (26.58) & $+47.13$ (26.64) & $+43.60$ (25.88) & $+0.73$ (1.72)\\
B & $-41.07$ (29.70) & $+56.53$ (29.55) & $+50.27$ (28.61) & $-0.60$ (2.11)\\
\bottomrule\end{tabular*}\end{table}

\begin{table}[htbp]\centering\footnotesize
\caption{\textbf{All fifteen fixed seed/template contrasts.}
Columns use the definitions in Table~\ref{tab:wording-summary}; $I_q$
additionally measures frozen-reader extractability. All values are pp.
The sign of seed $85$'s strict interaction varies even though its
boxed-to-hash payload change remains positive.}
\label{tab:wording-seed-contrasts}
\begin{tabular*}{0.97\linewidth}{@{\extracolsep{\fill}}llrrrrr@{}}
\toprule Template & Seed & $\Delta P_{B,h}$ & $I_P$ & $I_q$ & $I_S$ & $I_{\rm MV}$\\\midrule
Original & 83 & $-32.67$ & $+36.67$ & $+36.67$ & $+31.67$ & $-0.67$\\
Original & 84 & $-71.33$ & $+75.33$ & $+78.33$ & $+69.67$ & $-1.33$\\
Original & 85 & $+2.00$ & $+2.33$ & $-2.33$ & $-5.00$ & $-1.33$\\
Original & 86 & $-34.67$ & $+38.67$ & $+39.00$ & $+33.33$ & $0.00$\\
Original & 87 & $-71.00$ & $+75.33$ & $+76.00$ & $+69.67$ & $+2.00$\\
\midrule
A & 83 & $-44.67$ & $+57.67$ & $+57.67$ & $+52.33$ & $-0.33$\\
A & 84 & $-41.00$ & $+54.00$ & $+57.33$ & $+52.67$ & $+1.67$\\
A & 85 & $+11.00$ & $+1.67$ & $-1.00$ & $-1.33$ & $-0.67$\\
A & 86 & $-37.67$ & $+50.67$ & $+51.33$ & $+49.00$ & $+3.33$\\
A & 87 & $-59.00$ & $+71.67$ & $+72.00$ & $+65.33$ & $-0.33$\\
\midrule
B & 83 & $-52.67$ & $+68.00$ & $+67.33$ & $+57.33$ & $-2.00$\\
B & 84 & $-57.00$ & $+72.67$ & $+76.33$ & $+70.00$ & $+1.33$\\
B & 85 & $+10.33$ & $+5.33$ & $+5.00$ & $+1.67$ & $-2.67$\\
B & 86 & $-42.67$ & $+58.33$ & $+58.33$ & $+50.33$ & $-1.67$\\
B & 87 & $-63.33$ & $+78.33$ & $+78.33$ & $+72.00$ & $+2.00$\\
\bottomrule\end{tabular*}\end{table}

\paragraph{Scope.}
These are repeated measurements of five training pairs, not fifteen independent
training replications. Means are equally weighted across
seeds and SDs are sample SDs; no seed-population confidence interval
is inferred. The frozen analysis includes all $42/19800$ truncated responses
($0.21\%$); each cell has at most $3/300$ ($1.0\%$).
The check shows persistence under two fixed near-equivalent requests, not
generalization to a broad prompt distribution or semantic equivalence. Original-cache generation used different batch composition
from the new subset batches; matching per-item random seeds therefore
does not make differences between templates a deterministic prompt-only
effect. Within-template initial comparisons and their seed patterns
are the primary reported evidence. Machine-readable files include all
$66$ cell counts and rates, $15$ seed/template contrasts, frozen item
IDs, and exact prompt strings.

\subsection{MATH500 transfer: protocol and response diagnostics}
\label{app:transfer-preservation}
\label{app:math-transfer}

\begin{figure}[t]
\centering
\includegraphics[width=\textwidth]{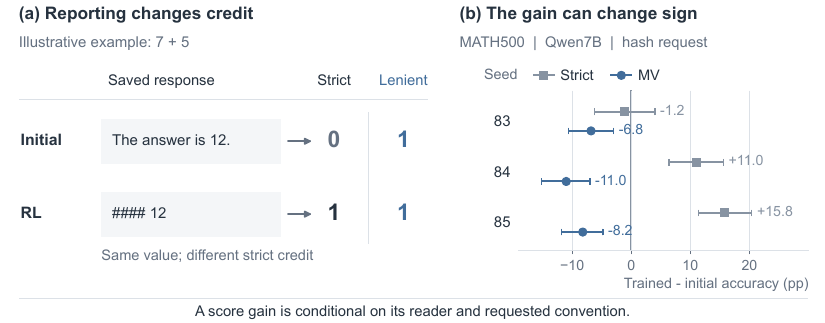}
\caption{\textbf{Reader dependence in an illustration and symbolic transfer.}
\textbf{(a)} An illustrative pair changes strict credit without changing
the displayed answer value; each response is scored by both readers.
\textbf{(b)} MATH500 hash-request gains for three Qwen2.5-7B training seeds: strict and Math-Verify (MV) score the same saved responses, with
95\% paired-item intervals conditional on each checkpoint. Two seeds
reverse gain direction between readers. Human calibration in
Table~\ref{tab:human-semantic-calibration} separately tests the strict gain
against committed-answer correctness; neither analysis validates chains.
The main controlled comparison appears in
Figure~\ref{fig:main} and Appendix Table~\ref{tab:five-seed-interactions}.}
\label{fig:math-transfer-overview}
\end{figure}

\paragraph{Prospective generation protocol.}
This section documents the original seed-$83$ crossed panel and its
diagnostics; additional hash-training seeds are reported separately in
Appendix~\ref{app:qwen7-seeds}. Table~\ref{tab:qwen7-math-transfer}
evaluates the Qwen2.5-7B initial checkpoint and the seed-$83$, step-$100$
GSM8K hash- and boxed-trained checkpoints on all $500$ items of
\texttt{HuggingFaceH4/MATH-500}. Each checkpoint receives both requested
conventions, producing $3{,}000$ responses.

Problem text is identical across requests; the suffixes are
``Let's think step by step and output the final answer after `\hash{}'.''
and
``Let's think step by step and output the final answer within
\boxedm{}.''
The original Qwen tokenizer and official chat template are held fixed
across checkpoints. Generation uses vLLM 0.11.0, Transformers 4.57.6,
bfloat16, $T=0.6$, top-$p=0.95$,
seed $20260909+\mathrm{item\ ID}$, an $8{,}192$-token output limit,
a $12{,}288$-token context limit, and batches of $32$. All prompts fit
without input truncation.

The output budget and a termination gate were frozen before test
generation. A $32$-item sample of MATH training problems, disjoint from
the test items and drawn with seed $20260909$, produced $64$
initial-model responses under the two requests. None truncated; the gate
allowed at most three truncations. The pilot was not selected by
correctness. Each raw response retains its item ID, seed, stop reason, and token
counts; all six formal cells and their archives passed integrity checks.

\begin{table}[!htb]
\centering
\caption{\textbf{MATH500 transfer: request-sensitive readout.}}
\label{tab:qwen7-math-transfer}

\footnotesize
\setlength{\tabcolsep}{3.2pt}
\renewcommand{\arraystretch}{1.08}

\begin{tabular*}{0.78\linewidth}{
@{\extracolsep{\fill}}
lrrrr
@{}}
\toprule

\rowcolor{TableBand}
\multicolumn{2}{@{}l}{\textbf{MATH500 transfer}}
&
\multicolumn{3}{r@{}}{\itshape seed $83$; $500$ items} \\

&
\multicolumn{2}{c}{Request \hash{}}
&
\multicolumn{2}{c}{Request \boxedm{}} \\
\cmidrule(lr){2-3}
\cmidrule(l){4-5}

Checkpoint & $S$ & MV & $S$ & MV \\
\midrule

Initial
& 53.0 & 73.4 & 75.0 & 75.4 \\

Hash-trained
& 51.8 & 66.6 & 76.6 & 76.8 \\

Boxed-trained
& \textbf{42.6} & \textbf{74.2} & 76.8 & 77.2 \\

\bottomrule
\end{tabular*}

\vspace{2pt}
\parbox{0.78\linewidth}{\scriptsize
Seed-$83$ GSM8K-trained checkpoints; values are accuracy (\%).
$S$: requested-convention symbolic reader; MV: full-response Math-Verify.
MV is not semantic ground truth. Protocol: Appendix~\ref{app:math-transfer}.}

\end{table}

\paragraph{Symbolic readers and conditional uncertainty.}
Strict hash scoring reads the first marker's same-line payload; strict
boxed scoring reads the first balanced \boxedm{} payload.
Math-Verify 0.9.0 checks each requested payload against the symbolic
reference with no numeric fallback. Full-response MV instead applies its
default candidate extraction to the same saved response. The two paths share symbolic-equivalence code; ``independent'' here refers to
extraction, not to an independent semantic oracle. 
All contrasts retain all $500$ items. We use $20{,}000$ shared paired
item-bootstrap resamples with seed $20260909$ and report percentile
$95\%$ intervals conditional on the fixed checkpoints and generations.
Neither a zero-containing interval nor a single training seed establishes
equivalence or training-seed stability.

Using the crossed interaction in Eq.~\ref{eq:crossinteraction}, the
seed-$83$ MATH500 training-convention $\times$ request interaction is
$+9.4$ strict points ($95\%$ CI $[3.6,15.4]$) and $-7.2$ MV points
($[-11.8,-2.6]$). Unlike the near-zero GSM8K MV interactions, the
MATH500 MV interaction changes materially under transfer. This result does not support a universally content-preserving interpretation
of convention learning.

\begin{figure}[!htb]
\centering
\includegraphics[width=\textwidth]{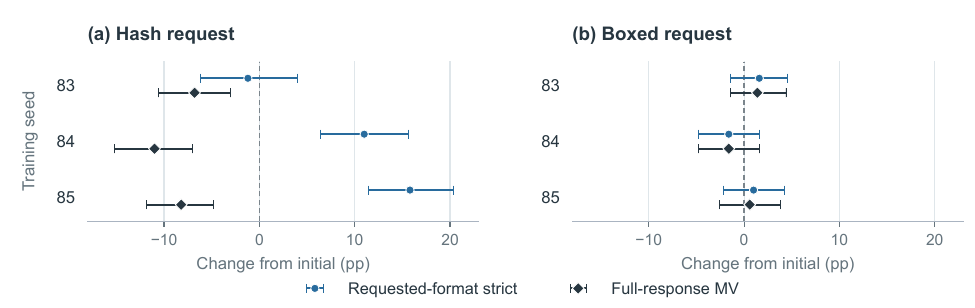}
\caption{\textbf{Transfer gains depend on both request and reader.}
GSM8K hash-trained Qwen2.5-7B checkpoints evaluated on the same $500$ MATH500
items. Changes are relative to the shared initial policy under each request. Points and bars show estimates and $95\%$ paired-item intervals
($20{,}000$ resamples), conditional on fixed checkpoints and responses.
Under hash requests, strict and MV changes have opposite signs in two
seeds; under boxed requests, all point estimates lie within $\pm2$ percentage points. These
automatic scores do not establish committed-answer gains.}
\label{fig:math-transfer-gains}
\end{figure}

\begin{table}[htbp]
\centering\footnotesize\setlength{\tabcolsep}{5pt}
\caption{\textbf{Full MATH500 transfer diagnostics.}
Percentages over $500$ items per cell; $q^{\rm pay}$ is presence of a nonempty
requested payload, $q$ its symbolic parseability, and Trunc.\ the
length-stop rate. $\Delta S$ and $\Delta\mathrm{MV}$ are changes in
percentage points from the same-request initial policy, with paired $95\%$ CIs.}
\label{tab:math-transfer-details}
\begin{tabular}{@{}llrrrll@{}}\toprule
Checkpoint & Request & $q^{\rm pay}$ & $q$ & Trunc. & $\Delta S$ [CI] & $\Delta\mathrm{MV}$ [CI]\\\midrule
Initial & hash & 63.4 & 60.0 & 0.6 & --- & ---\\
Hash-trained & hash & 99.4 & 98.2 & 1.0 & $-1.2$ [$-6.2$, $3.8$] & $-6.8$ [$-10.6$, $-3.2$]\\
Boxed-trained & hash & 48.6 & 47.8 & 0.6 & $-10.4$ [$-14.6$, $-6.2$] & $+0.8$ [$-2.4$, $4.0$]\\
\midrule
Initial & boxed & 99.0 & 98.8 & 0.4 & --- & ---\\
Hash-trained & boxed & 99.4 & 99.4 & 0.4 & $+1.6$ [$-1.4$, $4.6$] & $+1.4$ [$-1.6$, $4.4$]\\
Boxed-trained & boxed & 99.4 & 99.4 & 0.2 & $+1.8$ [$-1.2$, $4.8$] & $+1.8$ [$-1.2$, $4.8$]\\
\bottomrule\end{tabular}
\end{table}

\begin{table}[htbp]
\centering\footnotesize\setlength{\tabcolsep}{8pt}
\caption{\textbf{Exploratory answer-type diagnostics under hash requests.}
Exact counts; automatic grouping uses the reference's symbolic parse.
``Other'' includes all references not parsed as a nonnegative integer.
Digit-only means the hash payload matches an unsigned integer string.
This is a post-hoc reporting diagnostic, not a matched difficulty comparison.}
\label{tab:math-answer-types}
\begin{tabular}{@{}llrrr@{}}\toprule
Reference group & Checkpoint & Digit-only hash & Hash correct & MV correct\\\midrule
Nonnegative integer ($n=320$) & Initial & 191 & 178 & 246\\
 & Hash-trained & 314 & 246 & 245\\
 & Boxed-trained & 149 & 138 & 243\\\midrule
Other ($n=180$) & Initial & 3 & 87 & 121\\
 & Hash-trained & 159 & 13 & 88\\
 & Boxed-trained & 4 & 75 & 128\\
\bottomrule\end{tabular}
\end{table}

\paragraph{Post-hoc answer-report diagnostics.}
The primary analysis uses the prospectively frozen readers above. A later
offline diagnostic compares first and last requested markers in all saved responses and groups reference answers automatically according to whether
their symbolic parse is a nonnegative integer. This yields $320$ such
items and $180$ other items. The latter combine negative values,
fractions, symbolic expressions, complex values, and tuples rather than
forming a controlled difficulty stratum.

Under hash requests, initial/hash-trained/boxed-trained correct counts
remain $265/259/213$ when the last hash marker is read instead. Under
boxed requests, switching to the last boxed marker changes the corresponding
counts from $375/383/384$ to $376/384/386$. The main contrast is therefore
qualitatively insensitive to first-versus-last marker choice; this
exploratory diagnostic does not replace the primary scores.

Among hash-trained hash-request responses, $335$ contain both a
symbolically parseable hash payload and a symbolically parseable boxed
payload, of which $96$ are automatically judged unequal. The corresponding
unequal/both-parseable counts are $0/146$ for the initial checkpoint and
$2/230$ for the boxed-trained checkpoint. Because these automatic
disagreements can include notation failures, the $96$ cases are not
confirmed semantic contradictions.

There are also $67$ hash-trained responses with a correct boxed value but
an incorrect present hash value, compared with $3$ initially and $2$
after boxed training. The union of these $67$ cases and the $96$
automatic disagreements yields $99$ candidates for author review. This
candidate set has no separately reported human error breakdown; the
stratified aggregate human calibration is reported in
Appendix~\ref{app:human-calibration}. One response derives $-50$ for
$1-2+\cdots+99-100$, boxes $-50$, and then writes \hash{}~$50$;
another boxes $6+9i$ before writing \hash{}~$69$. Complete raw responses
are retained. A full-response reader can credit an earlier correct
expression even when the requested marker reports a conflicting value;
reader agreement is therefore not a substitute for labeling the
response's actual commitment.

Of the boxed-trained hash-request model's $89$ strict losses relative to
the initial checkpoint, $72$ remain correct under both boxed scoring and
MV. Overall, $16/3{,}000$ responses truncate. Removing any comparison
pair containing a truncation leaves the hash-trained hash-request MV
change at $-6.91$ points on $492$ pairs and the boxed-trained strict
change at $-10.53$ points on $494$ pairs. This is a post-hoc sensitivity
analysis rather than a replacement for the full-item comparison, and no
checkpoint was selected using these outcomes.

\section{Reader robustness, coverage, and evaluation protocol}
\label{app:reader-robustness}

\subsection{Matched evaluation and additional readers}
\label{app:matched}

\paragraph{Frozen protocol and actual decoding.}
The matched evaluation uses the retained initial and seed-$83$ step-$100$ RL
checkpoints for Qwen2.5-1.5B and SmolLM2, with identical per-family items and
prompts within each comparison. Sampled decoding uses seed
$20260908+\mathrm{item\ ID}$, $T=0.6$, top-$p=0.95$, and a
$640$-token output budget; greedy decoding uses $T=0$ and top-$p=1$. The context limit is $4{,}096$ tokens, with a pre-generation
length check and no silent input truncation. Responses are generated with
vLLM 0.11.0 and Transformers 4.57.6. Input hashes, model metadata, item
coverage, finish reasons, and token counts are retained.

All $32$ formal cells and four development cells completed; development
cells are excluded from scientific comparisons. The formal collection
contains $97{,}656$ responses: $31{,}656$ on GSM8K and $66{,}000$ on
the arithmetic probe.

\paragraph{Arithmetic construction and limits.}
Expressions use integers, parentheses, addition, subtraction,
multiplication, and exact integer division, with answers bounded by
$10{,}000$ in magnitude. The $500$ test items contain
$167/167/166$ two-/three-/four-operation expressions; the $32$
development fixtures are disjoint. Labels are checked by a separate
whitelisted AST evaluator using exact rational arithmetic. No test
response is used to select generation rules, prompts, or checkpoints.
This is a controlled arithmetic probe, not a distribution-free test of
reasoning; near-ceiling Qwen performance and possible pretraining overlap
limit its interpretation.

{\footnotesize
\setlength{\LTpre}{5pt}\setlength{\LTpost}{5pt}\setlength{\tabcolsep}{4pt}
\begin{longtable}{@{}llrrrrr@{}}
\caption{Complete matched GSM8K readouts (accuracy in percent), all $1319$ items per cell. $S_f/S_l$: first-/last-hash strict; $L_f$: hash-first/fallback; $L_n$: unconditional last number; MV: Math-Verify. These are scores on identical saved responses within each row.}\label{tab:matched-full}\\
\toprule
State & Prompt / $T$ & $S_f$ & $S_l$ & $L_f$ & $L_n$ & MV\\
\midrule\endfirsthead
\multicolumn{7}{l}{\emph{Table \ref{tab:matched-full}, continued}}\\
\toprule State & Prompt / $T$ & $S_f$ & $S_l$ & $L_f$ & $L_n$ & MV\\\midrule\endhead
\bottomrule\endfoot
\rowcolor{TableBand}\multicolumn{7}{@{}l}{\textbf{Qwen initial}}\\
 & Original / 0.6 & 11.07 & 11.07 & 70.74 & 70.89 & 71.42\\
 & Rephrased / 0.6 & 54.44 & 54.44 & 70.74 & 70.13 & 70.36\\
 & One-shot / 0.6 & 55.50 & 55.50 & 68.69 & 67.93 & 67.78\\
 & Original / 0 & 8.72 & 8.72 & 72.02 & 72.02 & 72.33\\
 & Rephrased / 0 & 57.85 & 57.85 & 70.58 & 69.83 & 69.83\\
 & One-shot / 0 & 56.41 & 56.41 & 69.14 & 68.16 & 68.31\\
\rowcolor{TableBand}\multicolumn{7}{@{}l}{\textbf{Qwen RL}}\\
 & Original / 0.6 & 75.36 & 75.36 & 75.36 & 75.28 & 75.28\\
 & Rephrased / 0.6 & 74.53 & 74.53 & 74.68 & 74.53 & 74.45\\
 & One-shot / 0.6 & 72.25 & 72.25 & 72.33 & 72.02 & 71.87\\
 & Original / 0 & 74.98 & 74.98 & 74.98 & 74.83 & 74.83\\
 & Rephrased / 0 & 74.83 & 74.83 & 74.91 & 74.53 & 74.45\\
 & One-shot / 0 & 74.00 & 74.00 & 74.45 & 74.37 & 74.30\\
\rowcolor{TableBand}\multicolumn{7}{@{}l}{\textbf{Smol initial}}\\
 & Original / 0.6 & 10.16 & 10.39 & 37.76 & 38.59 & 38.51\\
 & Rephrased / 0.6 & 36.92 & 36.92 & 41.17 & 41.17 & 41.09\\
 & One-shot / 0.6 & 39.73 & 39.73 & 42.46 & 42.38 & 42.38\\
 & Original / 0 & 12.05 & 12.43 & 40.64 & 41.17 & 41.09\\
 & Rephrased / 0 & 42.30 & 42.30 & 45.56 & 45.49 & 45.41\\
 & One-shot / 0 & 44.35 & 44.35 & 46.25 & 46.25 & 46.25\\
\rowcolor{TableBand}\multicolumn{7}{@{}l}{\textbf{Smol RL}}\\
 & Original / 0.6 & 41.47 & 41.55 & 42.91 & 42.61 & 42.61\\
 & Rephrased / 0.6 & 40.33 & 40.33 & 42.15 & 42.08 & 42.00\\
 & One-shot / 0.6 & 43.52 & 43.52 & 43.82 & 43.90 & 43.90\\
 & Original / 0 & 45.87 & 45.87 & 46.63 & 46.47 & 46.17\\
 & Rephrased / 0 & 44.96 & 44.96 & 46.25 & 46.25 & 46.25\\
 & One-shot / 0 & 45.19 & 45.19 & 45.19 & 45.19 & 45.19\\
\end{longtable}}

\begin{table}[ht]
\caption{\textbf{Fresh-arithmetic scores and measured generation cost.} $500$ items per state. Accuracy and pass@$32$ are percentages; output token totals count all $16000$ sampled responses. They describe generated work, not a latency or equal-compute claim.}
\label{tab:arithmetic-full}\centering\footnotesize
\setlength{\tabcolsep}{3.5pt}
\begin{tabular}{@{}llrrrrr@{}}
\toprule State & Readout & Strict & Fallback & MV & MV p@32 & Output tokens\\\midrule
Qwen initial & Greedy & 1.00 & 98.40 & 98.40 & --- & 54,350\\
Qwen initial & Sampled & 11.39 & 98.03 & 98.02 & 100.0 & 1,885,856\\
Qwen RL & Greedy & 98.80 & 98.80 & 98.80 & --- & 37,221\\
Qwen RL & Sampled & 98.42 & 98.42 & 98.42 & 100.0 & 1,202,385\\
Smol initial & Greedy & 18.40 & 62.20 & 77.80 & --- & 55,821\\
Smol initial & Sampled & 23.46 & 63.75 & 75.23 & 98.8 & 1,738,404\\
Smol RL & Greedy & 87.80 & 87.80 & 87.80 & --- & 45,160\\
Smol RL & Sampled & 77.50 & 80.58 & 81.12 & 99.0 & 1,658,272\\
\bottomrule
\end{tabular}
\end{table}
\begin{table}[ht]
\caption{Operation-stratified Math-Verify accuracy (percent), $32$ samples per item. Strata are defined before generation.}
\label{tab:arithmetic-strata}\centering\footnotesize
\begin{tabular}{@{}lrrr@{}}\toprule
State & Two operations ($167$) & Three ($167$) & Four ($166$)\\\midrule
Qwen initial & 99.64 & 98.75 & 95.67\\
Qwen RL & 99.48 & 98.56 & 97.21\\
Smol initial & 82.17 & 77.75 & 65.72\\
Smol RL & 91.69 & 83.29 & 68.32\\
\bottomrule
\end{tabular}
\end{table}

\paragraph{Additional matched OLMo evaluation.}
The OLMo initial model and completed RL checkpoints for seeds $83$--$85$ use
the same frozen OLMo prompt and all $1{,}319$ GSM8K items at both
$T=0.6$ and $T=0$, with the primary protocol's context limit, output
budget, and paired item seeds. This separate eight-cell collection adds
$10{,}552$ responses and is not pooled with either the primary
Qwen/Smol collection or any earlier OLMo panel. The seed-$84/85$
replications retain their step-$100$ checkpoints and raw distributed
shards.

Under sampled decoding, Math-Verify gains are positive in all three
training seeds, and all three paired intervals exclude zero
(Table~\ref{tab:olmo-matched}). For seeds $83/84$, the gains are
$+3.79$ and $+3.49$ points, with paired $95\%$ intervals
$[+1.82,+5.69]$ and $[+1.52,+5.53]$; seed $85$ supplies the third
positive result reported in the table and main matched analysis.
Under greedy decoding, lenient gains for seeds $83/84$ are
$+3.18$ $[+1.14,+5.08]$ and
$+1.82$ $[-0.08,+3.71]$, while the corresponding MV gains are
$+3.03$ $[+0.99,+4.93]$ and
$+1.67$ $[-0.23,+3.56]$. For seed $85$, the sampled lenient gain is
$+4.70$ points $[+2.81,+6.75]$ and the greedy gain is
$+1.90$ $[0.00,+3.79]$.

Thus the large strict--lenient gap reduction persists, while smaller
reader-robust gains can remain and their statistical support depends on reader
and decoding regime. This bounds a universal reporting-only interpretation. The three seed comparisons
share the same initial responses and therefore must not be treated as
independent item samples.

\begin{table}[ht]
\caption{\textbf{Matched OLMo-2-7B readouts for three completed seeds.}
All $1319$ GSM8K items, the same prompt, $640$ output tokens and $4096$
context tokens; scores are percentages. $S_f/S_l$: first-/last-hash strict;
$L_f$: hash-first/fallback; $L_n$: last number; MV: Math-Verify.}
\label{tab:olmo-matched}\centering\footnotesize
\setlength{\tabcolsep}{6pt}
\begin{tabular}{@{}llrrrrr@{}}
\toprule State & $T$ & $S_f$ & $S_l$ & $L_f$ & $L_n$ & MV\\\midrule
\rowcolor{TableBand}Initial & 0.6 & 65.58 & 65.58 & 81.35 & 81.20 & 81.58\\
RL seed 83 & 0.6 & 85.44 & 85.44 & 85.44 & 85.44 & 85.37\\
RL seed 84 & 0.6 & 85.06 & 85.06 & 85.06 & 84.99 & 85.06\\
RL seed 85 & 0.6 & 86.05 & 86.05 & 86.05 & 86.05 & 85.97\\
\rowcolor{TableBand}Initial & 0 & 66.72 & 66.72 & 82.94 & 82.87 & 82.94\\
RL seed 83 & 0 & 86.13 & 86.13 & 86.13 & 86.05 & 85.97\\
RL seed 84 & 0 & 84.76 & 84.76 & 84.76 & 84.76 & 84.61\\
RL seed 85 & 0 & 84.76 & 84.76 & 84.84 & 84.84 & 84.61\\
\bottomrule
\end{tabular}
\end{table}

\subsection{Independent re-scoring of the retained coverage panel}
\label{app:coverage-rescoring}

\paragraph{Frozen responses and re-scoring protocol.}
This analysis reuses the exact $1319\times256$ saved responses for each of
the initial and step-$100$ Qwen2.5-1.5B policies used in the coverage analysis of
\S\ref{sec:boundary}, with no new inference, response filtering, or
checkpoint selection. The original strict and hash-first/fallback
coverage curves are reproduced from the retained responses. We then
re-score the same responses with Math-Verify 0.9.0, SymPy 1.14.0, and
\texttt{latex2sympy2\_extended} 1.11.0 using default extraction and
numeric-reference normalization; unconditional last-number scoring
provides an additional reader check. Paired bootstrap intervals use
$2{,}000$ item resamples with seed $20260909$. This is a post-hoc robustness check on the frozen response collection.

\begin{table}[ht]
\centering
\footnotesize
\caption{\textbf{Coverage conclusions depend on the reader.}
Observed solved-set coverage at $k=256$ is reported in percent. Lost and
gained count item-level coverage changes from the initial to RL policy;
$p$-values are exact paired McNemar tests.}
\label{tab:p1}
\setlength{\tabcolsep}{5.5pt}
\renewcommand{\arraystretch}{1.05}
\begin{tabular}{@{}lrrrrr@{}}
\toprule
Reader & Initial & RL & Lost & Gained & $p$ \\
\midrule
First-hash strict    & $95.30$ & $98.64$ & $5$  & $49$ & $3.89\times10^{-10}$ \\
Hash-first/fallback  & $99.01$ & $98.64$ & $10$ & $5$  & $0.302$ \\
Last number          & $98.94$ & $98.64$ & $10$ & $6$  & $0.454$ \\
Math-Verify          & $99.17$ & $98.64$ & $10$ & $3$  & $0.092$ \\
\bottomrule
\end{tabular}
\end{table}

\paragraph{Reader-dependent coverage conclusions.}
At $k=256$, first-hash strict scoring yields $49$ gained and $5$ lost
items, producing a strong positive paired comparison
($p=3.89\times10^{-10}$). The same saved responses give $5$ gained and
$10$ lost items under hash-first/fallback scoring ($p=0.302$), $6$ gained
and $10$ lost under unconditional last-number scoring ($p=0.454$), and
$3$ gained and $10$ lost under Math-Verify ($p=0.092$). Thus the strong
positive coverage conclusion under the strict reader does not survive
the alternative readers. The latter comparisons establish neither a
coverage loss nor equivalence. As formalized by
Eq.~\ref{eq:support}, these quantities describe a random,
reader-dependent observed solved set under a finite sampling budget, not
the policy's full support.

\paragraph{Reader disagreements are substantive.}
The four initial-model items whose coverage verdict changes under the
additional reader checks are zero-based IDs $368$, $780$, $1042$, and
$1288$. On items $368$ and $1288$, a correct declared answer is followed
by a quantity copied from the question, causing unconditional last-number
scoring to miss the declared answer. On item $780$, a trailing constant
inside an unresolved expression receives last-number credit. On item
$1042$, MV extracts $3$ from responses whose final declared answer is
$-3.5$. These cases diagnose reader behavior rather than reasoning-chain
validity. We therefore retain the frozen readouts rather than selectively
correcting individual verdicts after inspection.

\subsection{Finite-budget coverage estimator and paired test}
\label{app:coverage-estimator}

The following standard estimator is used for the finite-budget coverage comparison in \S\ref{sec:design}.

\paragraph{Finite-budget coverage.}
Suppose item $i$ has $N$ retained responses, of which $c_i$ are credited
by a fixed reader. We estimate pass@$k$ using the standard unbiased
estimator \citep{chen2021evaluating},
\begin{equation}
\widehat{\mathrm{pass@}k}
=
\frac{1}{n}
\sum_{i=1}^{n}
\left[
1-
\frac{\binom{N-c_i}{k}}
     {\binom{N}{k}}
\right],
\qquad
1\le k\le N,
\label{eq:passk}
\end{equation}
with $\binom{N-c_i}{k}=0$ when $N-c_i<k$.

At $k=N$, this reduces to the fraction of items with at least one credited
response among the $N$ retained samples. For paired comparisons between
two policies, we classify each item as retained by both, lost
(initial-only), gained (trained-only), or solved by neither under the
same reader. Exact McNemar tests use only the paired lost/gained counts
to test equality of marginal success probabilities
\citep{mcnemar1947note}. A nonsignificant result is not an equivalence test,
and observed coverage at finite $k$ is not the full support of the policy.

\section{Training protocol, controls, and robustness}
\label{app:training-robustness}

\subsection{Training configuration and reward--readout distinction}
\label{app:train-config}

\paragraph{Matched 7B training configuration.}
The original matched pair uses full-parameter Qwen2.5-7B-Instruct,
GRPO, training seed $83$, $100$ updates, $32$ prompts per update, and
$8$ rollouts per prompt. AdamW uses a constant learning rate of
$10^{-6}$, weight decay $0.01$,
$(\beta_1,\beta_2)=(0.9,0.999)$, and gradient clipping at $1.0$.
Each update uses one policy epoch, minibatch size $32$, and microbatch
size $1$ per GPU, with symmetric policy clipping at $0.2$ and
sequence-mean/token-mean loss aggregation. Entropy loss, KL loss, and
KL reward shaping are disabled. GRPO normalizes advantages by the
within-group standard deviation.

Training rollouts use $T=1$, top-$p=1$, a maximum of $1024$ response
tokens, and a $3072$-token context; validation is greedy. Training uses
four GPUs, FSDP, bfloat16 computation, gradient checkpointing, and
parameter/optimizer offload. Token-level rollout importance correction
uses threshold $2$.

Additional hash-reward seeds $84/85$ use the same recipe, with only the
configuration changes documented in Appendix~\ref{app:qwen7-seeds}. The later boxed-reward seeds $84/85$
used for the paired GSM8K crossed replication follow the matched boxed
recipe described in Appendix~\ref{app:qwen7-crossed-seeds}. Both arms
are extended to seeds $86/87$ under the same paired recipe. These extensions leave the original seed-$83$ MATH500 crossed panel unchanged;
seeds $86/87$ add no MATH500 generations.

\paragraph{Frozen prompt proposal.}
Training prompts are sampled with replacement from three frozen
difficulty strata. Let $n_s$ denote the number of training items in
stratum $s$, and let $t$ and $v$ be normalized stratum-weight vectors
proportional to $(2302,3282,1912)$ and
$(24334,34693,40973)$, respectively. For item $i$ in stratum $s(i)$,
the target mass is
\[
P_i=\frac{t_{s(i)}}{n_{s(i)}},
\]
and the frozen proposal mass is
\[
Q_i=
(1-\varepsilon)
\frac{v_{s(i)}}{n_{s(i)}}
+
\varepsilon P_i,
\qquad
\varepsilon=10^{-6}.
\]
Each sampled prompt receives importance weight $P_i/Q_i$.

Proposal adaptation and stratified batching are disabled in these
completed configurations; consequently, the stored proposal exponent
$2$ does not induce online proposal updates. Strict target-preservation
checks and prompt importance correction are enabled in all matched arms.
This shared specialized design does not establish invariance to prompt
sampling. The estimator-robustness panel uses a common sampler setting across estimators
(Appendix~\ref{app:optimiser}).

\paragraph{Matched factors and convention-specific training rewards.}
The retained configuration/data check covers all $7{,}377$ training rows
and $1{,}319$ validation rows, verifying matching order and all
non-intervention fields. Convention-specific prompt suffixes and reward
extractors constitute the intervention; output paths necessarily differ.
This establishes matching of configured factors, not identical realized
policy trajectories.

The training reward searches only the final $300$ response characters for
the last match to the requested convention, removes commas and dollar
symbols, and compares the resulting payload exactly with the reference
target. The hash reward accepts a signed numeric payload following a space;
the boxed reward requires a closing brace, disallows nested braces, and
trims payload whitespace. Both return $1$ for an exact match and $0$
otherwise. These convention-specific extraction and syntax restrictions
are part of the training intervention.

The matched GSM8K evaluation uses a different readout: its strict reader
extracts the first numeric hash answer over the full response and compares
numerically with tolerance $10^{-3}$. Thus ``strict'' in the matched evaluation denotes an evaluation reader, not
the training reward program. For example, a response ending in
\texttt{\#\#\#\# 12.0} with reference \texttt{12} can receive strict
evaluation credit while failing the exact-string training reward.

This reward--readout distinction does not affect the within-reader
accounting identity in Eq.~\ref{eq:decomp}, because that identity compares
a fixed evaluation reader across checkpoints. It does, however, prevent
interpreting the strict evaluation gain as an exact change in training
reward. MATH500 transfer introduces symbolic evaluation readers only
after training; these checkpoints are not retrained with a symbolic
reward.

\paragraph{Post-hoc reward-program re-scoring.}
To test whether the crossed GSM8K result depends on the distinction above,
we apply each convention's actual training reward program to the saved
seed-$83$ crossed responses, with no new generation. The six cells contain $7{,}914$ responses in total. Training-program/evaluation-reader
correct counts are
$1013/1025$ for initial/hash,
$1165/1209$ for initial/boxed,
$1208/1208$ for hash-trained/hash,
$1216/1217$ for hash-trained/boxed,
$744/744$ for boxed-trained/hash, and
$1217/1217$ for boxed-trained/boxed.

Using the crossed contrast in Eq.~\ref{eq:crossinteraction}, the
training-convention $\times$ request interaction is $+35.25$ points
($95\%$ CI $[+32.37,+38.06]$) when responses are scored by the actual
training reward programs. Relative to the evaluation-readout interaction,
the change is only $+0.08$ points ($[-0.23,+0.45]$). These intervals use
$20{,}000$ shared paired item-bootstrap resamples with seed $20260910$;
the original seed-$83$ intervals reported in
Appendix~\ref{app:qwen7-crossed-seeds} remain unchanged. Removing the
$300$-character reward window changes no verdict in this retained sample.

Individual same-request gains are more sensitive to the scoring rule.
Under the boxed request, hash- and boxed-trained checkpoints gain
$+3.87$ and $+3.94$ points, respectively, under the training reward
programs, compared with $+0.61/+0.61$ points under the evaluation
readout. Thus training rewards and evaluation readers are not
interchangeable, while the seed-$83$ crossed interaction remains
essentially unchanged. This post-hoc re-scoring adds neither new generations nor independent semantic
labels.

\subsection{Controlled acquisition and reward/convention interventions}
\label{app:controlled-interventions}

\paragraph{Reward requirement control.}
For I1, the treatment keeps answer correctness as the reward target but removes the
requirement that the correct value appear after the requested marker. The control and
treatment use the same model family, task data, training budget, and evaluation
protocol within each comparison. Across three Qwen2.5-1.5B seeds, removing the
marker requirement reduces the mean strict gain from $+62.68$ to $+7.33$
percentage points ($88.3\%$ attenuation); across two SmolLM2 seeds, the
corresponding change is $+31.15$ to $-0.78$ points ($102.5\%$ attenuation).
The Qwen treatment varies substantially across seeds ($-10.41$ to $+21.93$
points), so the result supports strong attenuation rather than a deterministic
zero-gain effect.

\paragraph{Matched single-seed crossed reward control.}
\label{app:smol-reward-crossed}
A separate SmolLM2 seed-$83$ control holds the pretrained backbone, data order,
optimizer, schedule, and crossed evaluation fixed while changing whether the
reward requires the requested marker. Each evaluation cell contains $1{,}319$
unique IDs. The shared initial policy follows the hash and boxed requests on
$28.73\%$ and $17.21\%$ of items, respectively, so this panel starts from
low request adherence and is not directly comparable in magnitude with the
high-adherence boundary below.

\begin{center}
\footnotesize
\setlength{\tabcolsep}{5pt}
\renewcommand{\arraystretch}{1.05}
\begin{tabular}{lcc}
\toprule
Reward & Payload interaction & Strict interaction \\
\midrule
Marker-required & $+64.14$ & $+30.78$ \\
Answer-only & $-6.52$ & $-1.14$ \\
\midrule
Marker $-$ answer
& $+70.66\ [67.17,74.22]$
& $+31.92\ [28.43,35.48]$ \\
\bottomrule
\end{tabular}
\end{center}

Marker-required reward yields a $+64.14$-point requested-payload interaction,
whereas answer-only reward yields $-6.52$ points. Their paired difference is
$+70.66$ points with a $95\%$ item-bootstrap interval of
$[67.17,74.22]$; the corresponding strict-correct difference is $+31.92$
points $[28.43,35.48]$. Because this panel contains one training seed, these
intervals condition on the retained checkpoints and do not capture training-seed
uncertainty. The result therefore supports reward dependence of the crossed
format contrast, but does not establish that answer-only training restores
opposite-request adherence.

\paragraph{Requested/rewarded convention control.}
For I2, paired arms change both the requested training suffix and the
convention-specific reward extractor while holding the remaining training recipe
fixed. Qwen2.5-1.5B shows $+69.07$ and $+6.90$ point strict gains under the hash
and boxed conventions, respectively, corresponding to $90.0\%$ attenuation;
two additional paired seeds reproduce $89.93\%$ and $90.00\%$ attenuation.
SmolLM2 provides the planned boundary case: when neither convention is initially
well extracted, the mean strict gains are similar ($+29.64$ versus $+30.60$
points across two boxed-training seeds). These controls support a
familiarity-dependent reporting effect rather than a universal advantage of one
marker.

\subsubsection{Content-matched SFT: full protocol and cells}
\label{app:manufactured}

\paragraph{Content-matched supervision.}
Both SFT arms use the same $7{,}377$ GSM8K training rows. Each target
consists of the reference solution body followed by the arm-specific final
marker. The dataset builder verifies row by row that the two target
completions are byte-identical up to that marker and refuses to write a
pair otherwise; all $7{,}377$ pairs pass this check. The training prompt
contains only the question and no format instruction. The intervention therefore changes only the supervised reporting marker,
without explicitly requesting either convention during training. None of the $7{,}377$
training questions appears in the $500$-item evaluation panel.

\paragraph{Training and evaluation.}
Each arm is a full fine-tune of Qwen2.5-1.5B-Instruct for two epochs.
Seed $83$ uses learning rate $10^{-5}$ with cosine decay and $3\%$
warmup, effective batch size $32$, maximum sequence length $1024$, and
fp32 master weights under bfloat16 autocast, with one GPU per arm.
Final training losses are $0.1280$ for the \hash{} arm and $0.1284$ for
the boxed control arm. Seeds $84$ and $85$ use the identical recipe,
changing only the training seed.

Evaluation uses the same frozen request-swap harness used for the controlled convention evaluations: $500$ GSM8K items, four samples per item,
$T=0.6$, top-$p=0.95$, and the same frozen readers. Within each requested
convention, the two SFT arms use identical item files, verified by
\texttt{items\_sha256}
(\texttt{6ac44e28\ldots} for \hash{} and
\texttt{6243c83f\ldots} for \boxedm{}).

\begin{center}
\small
\begin{tabular}{llcccc}
\toprule
Model & Requested & Lenient & Strict & Gap & Strict recall \\
\midrule
Arm H (SFT \hash{})
  & \hash{}   & 0.5385 & 0.5380 & \textbf{0.0005} & 99.9\% \\
Arm H (SFT \hash{})
  & \boxedm{} & 0.5620 & 0.0000 & \textbf{0.5620} & 0.0\% \\
Arm B (SFT \boxedm{})
  & \hash{}   & 0.5615 & 0.0010 & \textbf{0.5605} & 0.2\% \\
Arm B (SFT \boxedm{})
  & \boxedm{} & 0.5560 & 0.5560 & \textbf{0.0000} & 100.0\% \\
\bottomrule
\end{tabular}
\end{center}

For seed $83$, the prompted convention preference of
Eq.~\ref{eq:dpref} is
\[
D(H)=+0.5380,\qquad
D(B)=-0.5550,
\]
compared with the frozen initial value
$D(\pi_0)=-0.5590$ in the frozen initial evaluation.
The arm-to-arm contrast is
\[
D(H)-D(B)=+1.0930,
\]
with a $95\%$ item-bootstrap interval
$[+1.0265,+1.1580]$ from $10{,}000$ resamples.

The prespecified criteria were all met: the \hash{}-marker arm
reverses the sign of $D$, the boxed control retains the initial sign, the
arm-to-arm contrast exceeds $+0.50$, lenient-score differences between
the two arms remain within $0.05$ for each request, off-convention gaps
substantially exceed on-convention gaps, and the \hash{} arm reverses the
initial model's preference. These checks constrain reporting behavior without assuming preservation of
answer quality.

Every value above is recomputed from per-item records by a readout script
independent of the experiment-driver log and cross-checked against the scorer
summary to $10^{-9}$. The two arms use disjoint output
directories, and the frozen protocol record predates the corresponding
metric files.

\paragraph{Replication across training seeds.}
Because the original cells use one training seed, item-level uncertainty
does not address training variability. Both SFT arms were therefore
retrained at seeds $84$ and $85$ under a separately frozen replication
protocol. The trainer, SFT corpora, evaluation item files, scorer, and
generation harness are unchanged; corpus hashes are verified at launch,
and the only varying training argument is the seed. Seed $83$ is not
regenerated or recomputed but is read from its original retained output.

\begin{center}
\small
\begin{tabular}{lllcccc}
\toprule
Seed & Model & Requested & Lenient & Strict & Gap & Strict recall \\
\midrule
$84$ & Arm H (SFT \hash{})
 & \hash{}   & 0.5385 & 0.5385 & \textbf{0.0000} & 100.0\% \\
$84$ & Arm H (SFT \hash{})
 & \boxedm{} & 0.5555 & 0.0000 & \textbf{0.5555} & 0.0\% \\
$84$ & Arm B (SFT \boxedm{})
 & \hash{}   & 0.5305 & 0.0000 & \textbf{0.5305} & 0.0\% \\
$84$ & Arm B (SFT \boxedm{})
 & \boxedm{} & 0.5530 & 0.5530 & \textbf{0.0000} & 100.0\% \\
\addlinespace[1pt]
$85$ & Arm H (SFT \hash{})
 & \hash{}   & 0.5295 & 0.5290 & \textbf{0.0005} & 99.9\% \\
$85$ & Arm H (SFT \hash{})
 & \boxedm{} & 0.5390 & 0.0000 & \textbf{0.5390} & 0.0\% \\
$85$ & Arm B (SFT \boxedm{})
 & \hash{}   & 0.5730 & 0.0000 & \textbf{0.5730} & 0.0\% \\
$85$ & Arm B (SFT \boxedm{})
 & \boxedm{} & 0.5540 & 0.5540 & \textbf{0.0000} & 100.0\% \\
\bottomrule
\end{tabular}
\end{center}

The arm-to-arm preference contrasts are
$+1.0915$ $[+1.0260,+1.1560]$ for seed $84$ and
$+1.0830$ $[+1.0180,+1.1470]$ for seed $85$, compared with
$+1.0930$ for seed $83$. Across all three seeds, the mean contrast is
$\mathbf{+1.0892}$ with a full spread of only $\mathbf{0.0100}$.

Averaging the convention preferences themselves gives
\[
D(H)=+0.5352,\qquad D(B)=-0.5540,
\]
compared with $D(\pi_0)=-0.5590$. Thus the final-marker intervention
reverses reporting preference in the \hash{} arm, whereas the
unchanged-marker control remains close to the initial preference.
The unchanged-marker control shifts by only $0.005$--$0.006$ in the two additional seeds,
while the marker-swapped arm changes the preference by about $1.09$.

All six replication criteria were met. In particular, both new seeds
replicate the sign reversal and exceed the prespecified $+0.50$
arm-to-arm contrast threshold; the cross-seed spread is $0.0100$ against
a $0.25$ ceiling. The validity criterion requiring request-wise
between-arm lenient-score differences no larger than $0.05$ also passes
for both seeds. Its closest margin is $0.0435$ in seed $85$, which we
report explicitly rather than rounding away.

These results show that final-marker supervision can install a strong
reporting preference under this controlled SFT design. They do
not establish preservation of answer quality: as reported in the main
text, both SFT arms have substantially lower lenient accuracy than the
initial model.

\subsection{Boundary cases}

\paragraph{Frozen OLMo-2 JSON/XML boundary.}
\label{app:olmo-jsonxml}
A screening protocol frozen before training selected JSON/XML as a familiar structured-output
pair: the initial policy followed the two requests on $251/256=98.05\%$ and
$250/256=97.66\%$ of screening items. We then train paired JSON- and XML-reward
arms on $7{,}377$ examples for $100$ steps at seeds $83$--$85$ and cross both
current requests against the shared initial policy. Each frozen evaluation cell
contains $1{,}319$ unique IDs; initial requested-payload presence is $97.42\%$
for JSON and $98.41\%$ for XML.

\begin{center}
\footnotesize
\setlength{\tabcolsep}{6pt}
\renewcommand{\arraystretch}{1.05}
\begin{tabular}{lcc}
\toprule
Seed & Payload interaction & Strict interaction \\
\midrule
$83$ & $+0.38$ & $-0.45$ \\
$84$ & $+0.30$ & $+1.74$ \\
$85$ & $+0.68$ & $+0.53$ \\
\midrule
Mean (seed SD) & $+0.45\ (0.20)$ & $+0.61\ (1.10)$ \\
\bottomrule
\end{tabular}
\end{center}

Requested-payload interactions remain small at all three seeds
($+0.38/+0.30/+0.68$ points), with mean $+0.45$ and seed SD $0.20$ points.
Under the opposite request, requested-payload presence does not deteriorate in
either trained arm; the observed changes relative to the shared initial policy
are positive in every seed. This is a negative boundary result: the large
hash/boxed interference does not automatically transfer to another model family
or to familiar structured-output interfaces. We do not use the exploratory
last-number readout from this panel as semantic evidence because it is neither
the paper's MV reader nor the human calibration.

\paragraph{Code transfer under an already familiar delimiter.}
\label{app:code}
On HumanEval+ and MBPP+ \citep{liu2023your,chen2021evaluating}, evaluation uses
execution-based unit tests and the relevant code-fence delimiter is already produced
reliably by the initial model. The initial formatting gaps are $0.00046$ and
$0.00001$, far smaller than the GSM8K hash-reader gap. Cross-domain transfer of
the RL gain is correspondingly small ($+0.0050/+0.0097$), while observed
pass@$64$ coverage changes by $-0.0180/-0.0148$. This is a contextual boundary,
not evidence that reader choice is generally irrelevant.

\subsubsection{Native-convention boundary on MATH}
\label{app:mathdecomp}

The GSM8K panels span substantial initial variation in
\hash{} extractability. MATH provides the complementary case in which
the task's familiar reporting convention is \boxedm{}, already produced
by Qwen2.5-1.5B-Instruct on $89.4\%$ of MATH500 items at step $0$.
Accordingly, the initial strict--lenient reader gap is only $0.0020$.

We train the native-\boxedm{} MATH arm for $100$ steps at three seeds
using the same model initialization and training recipe. Each retained
step-$100$ checkpoint is compared with the same frozen initial MATH500
responses. This panel uses its frozen MATH readers: strict
scoring requires an extractable \boxedm{} payload that is symbolically
equivalent to the reference, while the lenient path reads the
last line and otherwise falls back to the last number. These readers are
distinct from the later MATH500 transfer protocol in
Appendix~\ref{app:math-transfer}.

\begin{center}
\footnotesize
\setlength{\tabcolsep}{3pt}
\begin{tabular}{lccccccc}
\toprule
State
& Strict
& Lenient
& \boxedm{} extr.
& $\Delta_S$
& $\Delta_{\rm len}$
& $\gamma_{\rm RL}$
& Coverage $p$ (len./str.) \\
\midrule
Initial
& $0.528$ & $0.530$ & $0.894$
& --- & --- & --- & --- \\
RL seed $83$
& $0.550$ & $0.552$ & $0.892$
& $+0.022$ & $+0.022$ & $0.002$
& $0.774/1.0$ \\
RL seed $84$
& $0.542$ & $0.544$ & $0.886$
& $+0.014$ & $+0.014$ & $0.002$
& $0.804/0.607$ \\
RL seed $85$
& $0.528$ & $0.534$ & $0.888$
& $+0.000$ & $+0.004$ & $0.006$
& $0.629/0.629$ \\
\bottomrule
\end{tabular}
\end{center}

The initial reader gap is
\[
\gamma_0=0.530-0.528=0.0020
\]
for all three seed comparisons because the same initial responses are
reused. The resulting strict gains are $+0.022$, $+0.014$, and $0.000$,
while lenient gains are $+0.022$, $+0.014$, and $+0.004$.
Equation~\ref{eq:decomp} closes each comparison through the remaining
trained reader gaps $0.002$, $0.002$, and $0.006$.

For the two nonzero strict gains, the descriptive ratios of the initial
reader gap to the strict gain, $\gamma_0/\Delta_S$, are $9.1\%$ and $14.3\%$.
The ratio is undefined for seed $85$ because $\Delta_S=0$. As elsewhere,
these are descriptive ratios rather than causal shares. The present analysis
concerns the frozen MATH500 readout.

At $k=256$, none of the three paired observed-coverage comparisons is
significant under either frozen reader. For seed $83$, the
lenient-reader $p@256$ difference is $-0.004$ with interval
$[-0.018,+0.008]$. These zero-containing intervals do not establish
equivalence, and finite-budget coverage does not identify full policy
support.

The native-convention panel supplies a useful boundary case.
When the initial strict--lenient gap is only $0.0020$, strict gains are
small and closely track the corresponding lenient-reader changes:
$0.022/0.014/0.000$ under strict scoring versus
$0.022/0.014/0.004$ under the lenient reader. This is consistent with the broader observation that large
convention-sensitive gains are not inevitable when the initial reporting
interface is already aligned.

The comparison remains contextual, not causal: relative to the
GSM8K convention experiments, it changes the task, training split,
response distribution, and task-specific reader implementation in addition
to convention familiarity. It therefore bounds the interface-mismatch
account rather than isolating a task-independent effect of convention.

\subsection{Longer-horizon and estimator robustness}

\subsubsection{300-step horizon panel}
\label{app:horizon}
All primary RL interventions use a $100$-step budget. We therefore extend the
seed-$83$ Qwen2.5-1.5B hash control and boxed convention-swap arms to $300$ steps
under the same training recipe. The frozen validation protocol and all scheduled
checkpoints completed as intended.

\begin{center}
\footnotesize
\begin{tabular}{@{}lcccc@{}}
\toprule
Arm & Step $0$ & Step $100$ & Step $200$ & Step $300$ \\
\midrule
\hash{} control
& $0.0713$ & $0.7614$ & $0.7576$ & $0.7837$ \\
\boxedm{} swap
& $0.6876$ & $0.7579$ & $0.7799$ & $0.7713$ \\
\bottomrule
\end{tabular}
\end{center}

At $300$ steps the validation endpoints remain only $0.0124$ apart. Relative to
the hash control gain, the convention swap attenuates the strict validation gain by
$88.25\%$, close to the approximately $90\%$ attenuation at $100$ steps. On the
single-response matched readout, the lenient gain increases from $+0.0174$ at step
$100$ to $+0.0516$ at step $300$. Thus the convention-sensitive component remains
large at the longer horizon while reader-robust improvement can also grow. These
quantities are descriptive and do not support extrapolation beyond the observed
training window.

\subsubsection{Estimator robustness: balanced $2\times2$ panel}
\label{app:optimiser}
To test whether the convention-swap result depends on GRPO's group-relative
advantage estimator, we compare GRPO and Reinforce++
in a balanced $2\times2$ design with a common sampler setting. Within each
estimator, the requested/rewarded convention is the only experimental difference;
within each convention, the estimator is the only experimental difference.

\begin{table}[ht]
\caption{\textbf{Estimator $\times$ convention robustness panel.}
Qwen2.5-1.5B-Instruct, GSM8K, seed $83$, $100$ training steps, with
strict prompt sampling disabled in every cell. Values are mean
training-validation reward on the frozen $1{,}319$-item split.
No checkpoints were retained, so this panel does not support the
single-response strict/lenient decomposition used elsewhere. The two
\hash{} cells share the same step-$0$ value, as do the two \boxedm{}
cells, and all step-$0$ values reproduce the corresponding frozen
control runs exactly.}
\label{tab:optimiser}
\centering
\small
\setlength{\tabcolsep}{4pt}
\begin{tabular}{llccccc}
\toprule
Cell & Estimator & Convention & Step $0$ & Step $100$
& Gain & $\rho$ \\
\midrule
A & GRPO
  & \hash{}
  & $0.0713$ & $0.7544$ & $+0.6831$
  & \multirow{2}{*}{$\mathbf{0.8846}$} \\
B & GRPO
  & \boxedm{}
  & $0.6876$ & $0.7665$ & $+0.0788$
  & \\
\addlinespace[2pt]
C & Reinforce++
  & \hash{}
  & $0.0713$ & $0.7582$ & $+0.6869$
  & \multirow{2}{*}{$\mathbf{0.9040}$} \\
D & Reinforce++
  & \boxedm{}
  & $0.6876$ & $0.7536$ & $+0.0660$
  & \\
\bottomrule
\end{tabular}
\end{table}

Using Eq.~\ref{eq:rho}, attenuation is $88.46\%$ for GRPO and $90.40\%$ for
Reinforce++, an absolute difference of $1.94$ percentage points. Both convention
arms finish close to one another within each estimator. This panel covers one model,
task, seed, training budget, and sampler setting; it therefore supports estimator
robustness within this frame rather than invariance across RL algorithms generally.

\section{Extended related work}
\label{app:related}

\paragraph{Positioning and scope.}
Format adaptation, reader errors, and instruction-following losses are
established starting points (Table~\ref{tab:novelty-map}). Our contribution separates the evidence needed for three claims about an
update:
whether it acquires a reporting convention, follows the current request,
and improves committed-answer correctness. The content-matched final-marker
SFT intervention tests whether reporting preference itself is learnable;
the RL comparison crosses training convention with the current request and
then reads the resulting responses under multiple evaluators; human
calibration compares selected reader-measured gains with committed-answer
correctness in its sampled settings. The human calibration does not cover
the complete 7B crossed-request matrix. A static reader comparison alone
does not establish learned reporting behavior, while behavioral controls
alone do not establish semantic improvement. These experiments constrain distinct interpretations without sharing a single
causal estimand.

\begin{table}[htbp]
\caption{Related questions and the evidence added by the present study.
Distinctions describe designs, not priority claims.}
\label{tab:novelty-map}\centering\footnotesize
\setlength{\tabcolsep}{4pt}
\begin{tabular}{@{}>{\raggedright\arraybackslash}p{.22\textwidth}>{\raggedright\arraybackslash}p{.34\textwidth}>{\raggedright\arraybackslash}p{.39\textwidth}@{}}
\toprule
Related work & Finding already studied & Distinction tested here\\
\midrule
The Invisible Leash \citep{wu2025invisible} &
Finite-budget support, including prompt-format misalignment and
revised answer processing. &
Acquisition of a convention versus following a new request;
the crossed comparison pairs both trained arms within each seed.\\
\addlinespace[3pt]
From Accuracy to Robustness \citep{huang2025accuracy} &
Verifier classification errors and behavior under RL training,
including vulnerability to reward hacking. &
Whether a request-dependent training advantage persists under
another reader; selected gain contrasts receive a human semantic check.\\
\addlinespace[3pt]
Format correction / imitation \citep{wang2026reinforcement,yang2025emperor} &
Gains beyond format correction and effects of SFT format imitation. &
Final-marker preference under exactly matched target content;
acquisition remains distinct from the observed answer-score losses.\\
\addlinespace[3pt]
MathIF \citep{fu2026scaling} &
Instruction following and reasoning-oriented post-training. &
Training convention crossed with request and reader; marker absence
is distinguished from marker presence with conflicting extracted values.\\
\bottomrule
\end{tabular}
\end{table}

\paragraph{Format learning and weak supervision.}
SFT format imitation \citep{yang2025emperor} concerns worked-solution
structure; our content-matched SFT contrast changes only the final marker,
directly testing the learnability of reporting preference without assuming
preserved answer quality or capability. Parseable-answer rewards can
also improve accuracy beyond format correction \citep{wang2026reinforcement}.
Spurious-reward mechanisms \citep{shao2025spurious,chen2026exploration} and input-template
effects \citep{liu2025understanding} require matched tests before attributing their
gains to the extraction gap studied here.

\paragraph{Instruction following and value preservation.}
General instruction-following benchmarks evaluate compliance with verifiable
or fine-grained constraints \citep{zhou2023instruction,jiang2024followbench}.
MathIF \citep{fu2026scaling} studies instruction following during
mathematical reasoning and its tradeoff with reasoning-oriented
post-training. Our training-convention $\times$ current-request comparison addresses a
narrower reporting question: whether a learned reporting
preference competes with the convention requested at evaluation. Symbolic
transfer further distinguishes marker absence from cases in which the
requested marker is present but carries a value that conflicts with an
earlier answer. 
\paragraph{Coverage, reasoning validity, and semantic correctness.}
Answer-based coverage \citep{chen2026does}, chain correctness
\citep{wen2026reinforcement}, and contamination \citep{wu2026reasoning}
address different sources of apparent improvement. Our finite-budget
coverage analysis remains reader-dependent and does not identify full
policy support. Causal importance and sufficiency of reasoning
\citep{yu2026outcome} require chain-level tests that our answer-level
audits do not perform, while human calibration targets committed-answer
correctness rather than derivation validity. Re-scoring alone cannot distinguish valid reasoning from memorization or lucky
answers.

\paragraph{Evaluation and training context.}
Prompt sensitivity
\citep{sclar2024quantifying,alzahrani2024benchmarks,mizrahi2024state,hochlehnert2025sober}
motivates these controls. Reward gaming and reward-model overoptimization
\citep{skalse2022defining,gao2023scaling,rafailov2024scaling} limit inference
from reward alone, but correct formatting need not constitute reward hacking.
Token-level analyses reveal heterogeneous optimization effects across tokens
\citep{wang2026beyond}, which our output-level evidence does not localize.
Our setting builds on chain-of-thought
prompting, instruction tuning, and PPO
\citep{wei2022chain,ouyang2022training,schulman2017proximal}, with standard
rollout infrastructure \citep{sheng2025hybridflow,kwon2023efficient}.
Selected prospective controls use frozen protocols; the appendix focuses on analyses that directly bear on the paper's stated claims.

\end{document}